%% file: main.tex
\documentclass{article}
\usepackage{microtype}
\usepackage{amsmath,amssymb,amsfonts}
\usepackage{graphicx}
\usepackage{textcomp}
\usepackage{xcolor}
\usepackage{tikz}
\usetikzlibrary{arrows.meta,backgrounds,calc,fit,positioning}
\usepackage{booktabs}
\usepackage{array}
\usepackage{multirow}
\usepackage{xspace}
\usepackage[hyphens]{url}
\usepackage{placeins}
\usepackage[caption=false,font=footnotesize]{subfig} %
\usepackage{hyperref}

\usepackage[accepted]{preprint}

\usepackage{cleveref}

\newcommand{\sys}{FOSCAR\xspace} %

\definecolor{ptblue}{HTML}{4477AA}
\definecolor{ptcyan}{HTML}{66CCEE}
\definecolor{ptgreen}{HTML}{228833}
\definecolor{ptyellow}{HTML}{CCBB44}
\definecolor{ptred}{HTML}{EE6677}
\definecolor{ptpurple}{HTML}{AA3377}
\definecolor{ptgrey}{HTML}{BBBBBB}

\Crefname{figure}{Fig.}{Figs.}
\Crefname{section}{Sec.}{Secs.}
\Crefname{algorithm}{Algo.}{Algos.}
\Crefname{equation}{Eq.}{Eqs.}

\papertitlerunning{FOSCAR}

\begin{document}

\twocolumn[
\papertitle{FOSCAR: Accelerating Decode Attention over 2- and 4-Bit KV Caches through Affine Folding}

\begin{paperauthorlist}
\paperauthor{Donglin Zhuang}{together,usyd}
\paperauthor{Pragaash Ponnusamy}{together}
\paperauthor{Xiaoxia Wu}{together}
\paperauthor{Shuaiwen Leon Song}{usyd}
\paperauthor{Zhongzhu Zhou}{together,usyd}
\end{paperauthorlist}

\paperaffiliation{together}{Together AI}
\paperaffiliation{usyd}{University of Sydney}

\papercorrespondingauthor{Donglin Zhuang}{donglinzhuang@outlook.com}

\paperkeywords{KV cache quantization, decode attention, GPU kernels, Tensor Cores}

\vskip 0.3in

\input{src/abstract}
]

\printAffiliationsAndNotice{}

\input{src/0-intro}
\input{src/1-background}
\input{src/2-analysis}
\input{src/3-design}
\input{src/4-serving}
\input{src/5-evaluation}
\input{src/6-related}
\input{src/7-conclusion}

\FloatBarrier
\bibliography{refs}
\bibliographystyle{preprint}

\clearpage
\label{appendix-start}
\input{src/8-appendix}

\end{document}

%% file: src/abstract.tex
\begin{abstract}
Large-language-model decode attention streams the key--value (KV) cache for every generated token.
Packing that cache into 2-bit or 4-bit quantized integers reduces memory traffic but does not automatically translate into latency saving.
Existing 2-bit fused kernels only reach less than $31\%$/$15\%$ of peak HBM throughput on H100/B200.
We identify that converting each value and applying its scale and offset before feeding Tensor Cores is a major bottleneck because instruction-issue throughput is a scarce resource on modern GPUs.
\sys{} does not apply scale and offset to each value from the KV cache, but uses \emph{affine folding} to apply scale and offset to either the other operand or the output of matrix products in decode attention, depending on the quantization scheme (token-wise or channel-wise).
This reduces scale-and-offset operations by up to two orders of magnitude while preserving the dequantized attention output in exact arithmetic.
\sys{} includes carefully designed bit manipulation to feed KV operands to Tensor Cores efficiently, requiring only $1$ and $0.19$ ALU instructions per element for the INT2 KV cache on H100 and B200, respectively.
Finally, \sys{} builds warp-specialized GPU pipelines for Hopper (H100) and Blackwell (B200), leveraging architecture-specific features such as \texttt{wgmma} and \texttt{tcgen05}.
On H100, \sys{} INT2 and INT4 kernels achieve geo-mean speedups of $3.36\times$ and $2.83\times$ over FlashInfer BF16, and $2.03\times$ and $1.71\times$ over TensorRT-LLM FP8.
On B200, the corresponding INT2 and INT4 speedups are $2.67\times$ and $2.30\times$ over FlashInfer BF16, $1.48\times$ and $1.28\times$ over TensorRT-LLM FP8, and $1.16\times$ and $1.00\times$ over TensorRT-LLM NVFP4.
We integrate \sys{} into SGLang to benchmark serving efficiency.
On long-context serving workloads, \sys{} increases peak aggregate throughput by up to $170\%$ and $341\%$ over 2-bit and 16-bit baselines, respectively.
We open-source \sys{} as a kernel library at \url{https://github.com/FutureMLS-Lab/FOSCAR}.
\end{abstract}

%% file: src/0-intro.tex
\section{Introduction}
\label{sec:intro}

Autoregressive language-model decoding reads each active request's key--value (KV) cache on every decode step.
The cache footprint grows with batch size and context length; storing it consumes high-bandwidth memory (HBM) capacity, and streaming it from HBM often dominates decode-attention latency.
Low-bit KV quantization reduces that traffic directly.
KIVI~\cite{liu2024kivi}, Kitty~\cite{xia2025kitty}, OSCAR~\cite{zhou2026oscar}, KVQuant~\cite{hooper2024kvquant}, and QServe~\cite{lin2025qserve} store each cached value in 2--4 bits with a group-shared scale and offset.

Compressing the cache in memory does not eliminate on-chip operand preparation.
KIVI~\cite{liu2024kivi}, QServe~\cite{lin2025qserve}, TurboAttention~\cite{kang2025turboattention}, BitDecoding~\cite{du2026bitdecoding}, SAW-INT4~\cite{jia2026saw}, and OSCAR~\cite{zhou2026oscar} all convert or reconstruct low-bit values before each matrix product.
This preparation consumes instruction-issue throughput, registers, and shared-memory bandwidth even when fused into attention without writing dequantized K and V to HBM.
\emph{Although bit-packing reduces HBM traffic, per-element operand preparation can still limit the resulting latency speedup.}

\sys{} applies affine folding to the query--key (QK) and probability--value (PV) matrix products, which reduce over channels and tokens.
For token-wise K, the kernel scales each group's partial QK result and adds the group's offset times the matching query sum.
For token-wise V, the kernel scales the attention weights before PV and adds a weighted sum of offsets to the output.
Channel-wise K instead scales the query before QK, while channel-wise V scales each group's partial PV result, and both add the corresponding offset-weighted sum.
These per-group operations replace element-wise scale multiplication and offset addition, while bit manipulation constructs Tensor Core KV operands directly from packed integers.
In exact arithmetic, affine folding produces the same attention output as dequantizing every KV value before the matrix products.

This paper makes four contributions:
\begin{itemize}
  \item \textbf{Identify decode attention performance bottleneck.}
        We compare per-byte reconstruction instruction demand against peak instruction-issue and HBM throughputs on H100 and B200, showing how per-element conversion and affine arithmetic limit low-bit decode speedups (\Cref{sec:analysis}).
   \item \textbf{Efficient Tensor Core operand preparation via affine folding.}
        We formulate affine folding for independent token-wise or channel-wise K and V quantization, paired with bit-level Tensor Core operand construction, to feed Tensor Cores efficiently with lower instruction demand (\Cref{sec:design-fold,sec:design-unpack}).
  \item \textbf{Fast GPU kernels for Hopper and Blackwell.}
        We design warp-specialized decode pipelines for H100 register accumulators and B200 tensor-memory (TMEM) accumulators (\Cref{sec:design-pipelines}).
  \item \textbf{Kernel and serving-engine evaluation.}
        We evaluate \sys{} in standalone kernel benchmarks and integrated into SGLang~\cite{zheng2024sglang}.
        \sys{} INT2 achieves geometric-mean kernel-latency speedups over FlashInfer BF16 of $3.36\times$ on H100 and $2.67\times$ on B200.
        For long-context serving workloads, \sys{} INT2 increases peak aggregate serving throughput by up to $170\%$ and $341\%$ over 2-bit and 16-bit baselines, respectively (\Cref{sec:eval}).
\end{itemize}

%% file: src/1-background.tex
\section{Background and Motivation}
\label{sec:background}

\subsection{Decode attention with a quantized KV cache}
\label{sec:bg-quant}

Each decode step streams the cached K and V tensors from HBM, so low-bit storage reduces the bytes that attention must read.
Asymmetric quantization stores each cached element as a $b$-bit quantized value $q\in\{0,\ldots,2^b-1\}$.
Each quantization group shares a scale $s$ and an additive offset $z$ to reconstruct an element $x$ as
\begin{equation}
x=sq+z.
\label{eq:dequant}
\end{equation}
For either K or V, a token-wise group shares a scale and offset across channels within one token~\cite{yao2022zeroquant}, whereas a channel-wise group shares them across tokens within one channel~\cite{banner2019posttraining}.

\subsection{Limited latency savings in low-bit decode}
\label{sec:bg-latency}

Low-bit storage reduces the bytes that an attention kernel reads from HBM, but existing kernels still convert and dequantize each cached value before the matrix product.
In \Cref{fig:roofline-gap}, two- and four-bit kernels achieve a much smaller fraction of peak HBM throughput on their KV traffic than BF16 kernels, whether they fuse dequantization into attention~\cite{lin2025qserve,kang2025turboattention,du2026bitdecoding,jia2026saw,zhou2026oscar} or into separate matrix-vector kernels~\cite{liu2024kivi}.

Unpacking each integer value and applying its group scale and offset consume instruction-issue slots.
As the stored bit width drops, HBM transfer time falls while this per-value instruction cost stays fixed, shifting the bottleneck from memory bandwidth to instruction issue.
\Cref{sec:analysis} quantifies this instruction-issue cost and shows how affine folding removes per-value affine arithmetic.

%% file: src/2-analysis.tex
\section{Reconstruction limits low-bit decode attention speedups}
\label{sec:analysis}

A low-bit KV cache reduces HBM traffic, but the kernel must still construct floating-point operands fast enough to supply the Tensor Cores.
In existing baselines, reconstruction involves converting integer values into floating-point operands, applying the affine transformation $sq+z$, and optionally performing one additional conversion to match the Tensor Core operand format (e.g., BF16 to FP8).

\subsection{Low-bit baselines fall short of peak HBM throughput}
\label{sec:analysis-gap}

\input{figures/roofline-gap}

We benchmark FlashInfer~\cite{ye2025flashinfer}, TensorRT-LLM~\cite{nvidia2026trtllm}, FlashAttention-2~\cite{dao2024flashattention2}, FlashAttention-3~\cite{shah2024flashattention3}, FlashAttention-4~\cite{zadouri2026flashattention4}, BitDecoding~\cite{du2026bitdecoding}, KIVI~\cite{liu2024kivi}, Kitty~\cite{xia2025kitty}, SAW-INT4~\cite{jia2026saw}, and OSCAR~\cite{zhou2026oscar} across KV-cache storage formats.
We compute effective KV access throughput by dividing the bytes in one read of the stored values, scales, and offsets by decode time.
\Cref{fig:roofline-gap} expresses this rate as a fraction of each GPU's peak HBM throughput at batch size $B=128$ and context length $KV=32\mathrm{K}$.

FlashInfer BF16 exceeds $90\%$ of peak on both H100 and B200.
On B200, hardware-native TensorRT-LLM NVFP4 reaches $72.4\%$ of peak in the 4-bit group.
Among integer 4-bit baselines, BitDecoding is highest at $70.0\%$ on H100 and $36.1\%$ on B200, while its INT2 kernel reaches only $30.7\%$ and $14.5\%$, respectively.
\sys{} reaches $91.3\%$ and $60.6\%$ at INT4 and INT2 on H100, and $68.6\%$ and $44.4\%$ on B200.
Although low-bit baselines read fewer KV-cache bytes, their effective KV access throughput remains far below peak because kernel latency is bounded by the slowest pipeline stage.
\Cref{sec:analysis-issuerate} shows how operand reconstruction can demand more instruction-issue throughput than the GPU provides at peak HBM throughput.

\subsection{Reconstruction consumes instruction-issue slots}
\label{sec:analysis-issuerate}

HBM transfers, operand preparation, softmax, and Tensor Core products all require instructions from the warp schedulers.

\input{figures/reconstruction-demand}

From H100 to B200, peak HBM throughput grows $2.39\times$, while peak instruction-issue throughput grows only $11\%$~\cite{nvidia2025h100spec,nvidia2025dgxb200,nvidia2025hgxb200,nvidia2026cudabpg}.
Peak dense Tensor Core throughput grows $4.55\times$ when comparing H100 FP8 with B200 FP4~\cite{nvidia2025h100spec,nvidia2025dgxb200,nvidia2025hgxb200}.
\Cref{fig:machine-balance} compares these peak throughputs across GPU generations\footnote{Rubin's figures are preliminary; we infer its instruction-issue throughput from FP32 CUDA Core throughput assuming 128 lane-slots per SM cycle~\cite{nvidia2023gpuperformance,nvidia2026rubin,nvidia2026cudabpg}.}.

Narrower stored values reduce HBM traffic without proportionally reducing reconstruction work.
We count the instructions to convert packed integers to BF16, evaluate $sq+z$ in BF16, and convert the results to FP8 or FP4 when the Tensor Core instructions require those formats.
One lane-slot corresponds to one instruction issued for one warp lane~\cite{nvidia2026cudaguide}.
Let $a_{\mathrm{issue}}$ denote the lane-slots required to construct an operand and evaluate $sq+z$ for one quantized value, $C_{\mathrm{issue}}$ the GPU's peak instruction-issue throughput in lane-slots/s, and $B_{\mathrm{HBM}}$ its peak HBM throughput in bytes/s.
Each quantized value occupies $B_{\mathrm{elem}}=0.5$~bytes in INT4 and $0.25$~bytes in INT2, excluding scales and offsets.
Streaming the cache at peak HBM throughput remains within peak instruction-issue throughput only when
\begin{equation}
  \underbrace{\frac{a_{\mathrm{issue}}}{B_{\mathrm{elem}}}}_{\text{issue demand per byte}}
  \leq
  \underbrace{\frac{C_{\mathrm{issue}}}{B_{\mathrm{HBM}}}}_{\text{peak issue per byte}}
  \label{eq:issue-balance}
\end{equation}
\Cref{fig:ops-per-byte} compares this per-byte issue demand against the hardware peak from \Cref{eq:issue-balance} on H100 and B200.

Preparing four BF16 Tensor Core operands on H100 requires two \texttt{SHF} and two \texttt{LOP3} instructions for INT$\to$FP conversion.
Two BF16 \texttt{HFMA2} instructions then evaluate the affine transformation $sq+z$, each producing two BF16 values in a 32-bit register~\cite{frantar2025marlin}.
The total is six instructions, or six lane-slots per warp lane.
Four values occupy two bytes in INT4 or one byte in INT2, giving $6/2=3$ or $6/1=6$ lane-slots/byte in~\Cref{fig:ops-per-byte}, respectively.
Therefore, reconstruction alone can consume $30\%$ and $60\%$ of H100's peak hardware budget of $9.99$ lane-slots/byte, leaving little room for other parts of the kernel.

On B200, two \texttt{F2FP} instructions convert the four BF16 values directly to FP8 or FP4 and pack them into registers.\footnote{Before PTX ISA 9.1~\cite{nvidia2026ptx}, the conversion uses FP32 intermediates and takes six instructions per four values with CUDA 13.0.
Reconstruction then costs twelve instructions, or $6$ and $12$ lane-slots/byte for INT4 and INT2, respectively.}
Reconstructing four values therefore takes $6+2=8$ lane-slots, or $8/2=4$ lane-slots/byte for INT4 and $8/1=8$ lane-slots/byte for INT2.
At peak HBM throughput, INT2 reconstruction \emph{exceeds} B200's hardware peak of $4.65$ lane-slots/byte, and INT4 reconstruction alone consumes $86\%$ of that peak.

\sys{} removes per-value affine evaluation and never materializes a reconstructed K or V tile.
Instead, it constructs Tensor Core operands directly from $q$, applies each group's scale to the other matrix input or partial product, and adds the offset as an output correction.
At four query heads per KV head and group size $128$, affine folding reduces issue demand in \Cref{fig:ops-per-byte} to $2.13$ and $4.25$ lane-slots/byte (INT4 and INT2) on H100, and to $0.88$ and $1.00$ on B200.

%% file: figures/roofline-gap.tex
\begin{figure}[t]
  \centering
  \captionsetup[subfloat]{farskip=2pt}
  \includegraphics[width=\columnwidth]{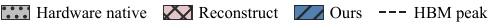}
  \par
  \subfloat[H100 (3.35~TB/s peak).\label{fig:roofline-gap-h100}]{%
    \includegraphics{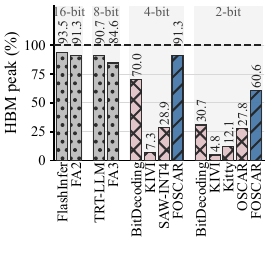}}
  \hfill
  \subfloat[B200 (8.00~TB/s peak).\label{fig:roofline-gap-b200}]{%
    \includegraphics{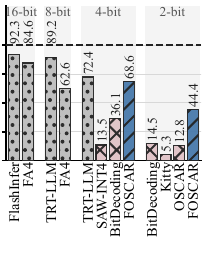}}
  \caption{Effective KV access throughput of baseline kernels and \sys{} at $B=128$ and $KV=32\mathrm{K}$ as a percentage of peak HBM throughput.}
  \label{fig:roofline-gap}
\end{figure}

%% file: figures/reconstruction-demand.tex
\begin{figure}[t]
  \centering
  \subfloat[Peak throughput across GPU generations, normalized to H100 (labels show Rubin).\label{fig:machine-balance}]{%
    \includegraphics[width=0.49\columnwidth]{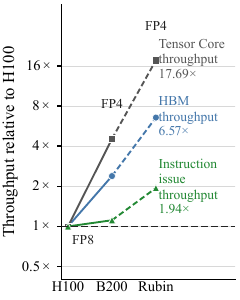}}
  \hfill
  \subfloat[Instruction-issue demand per packed KV byte on H100 and B200. INT$\to$FP for converting packed integers to floating-point values; Affine is for apply scale and offset.\label{fig:ops-per-byte}]{%
    \includegraphics[width=0.49\columnwidth]{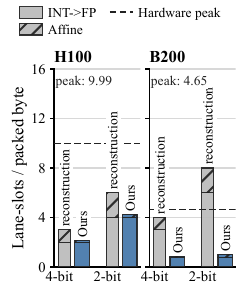}}
  \caption{GPU throughput scaling and per-byte instruction-issue demand for operand construction and affine evaluation (\sys{} shown at four query heads per KV head and group size $128$).
  }
  \label{fig:reconstruction-demand}
\end{figure}

%% file: src/3-design.tex
\section{\sys{} Design}
\label{sec:design}

\sys{} consists of three components:
\begin{itemize}
  \item \emph{Affine folding} moves shared scales and offsets from individual KV values to matrix-product inputs and results so that Tensor Cores consume operands built directly from packed quantized values, reducing the affine demand in \Cref{fig:ops-per-byte} (\Cref{sec:design-fold}).
  \item \emph{Bit-level Tensor Core operand construction} builds Tensor Core operands directly from packed quantized values, reducing the INT$\to$FP conversion demand in \Cref{fig:ops-per-byte} (\Cref{sec:design-unpack}).
  \item \emph{Warp-specialized GPU pipelines} overlap data movement, operand preparation, matrix products, and affine corrections on Hopper and Blackwell (\Cref{sec:design-pipelines}). 
\end{itemize}

\subsection{The affine folding mechanism}
\label{sec:design-fold}

\input{figures/fig_fold_mechanism}
Consider $C=A(sq+z)$, where $A\in\mathbb{R}^{M\times K}$, $q\in\mathbb{R}^{K\times N}$ holds quantized values, and $k$ indexes the reduction.
When each row $k$ of $q$ shares a scale $s_k$ and offset $z_k$ across columns $j$, distributing reconstruction through the product gives
\begin{equation}
  \begin{aligned}
    C_{i,j} &= \sum_k A_{i,k}(s_kq_{k,j}+z_k) \\
      &= \underbrace{\sum_k(s_kA_{i,k})q_{k,j}}_{\text{scaled input}}+
         \underbrace{\sum_k z_kA_{i,k}}_{\text{output correction}}.
  \end{aligned}
  \label{eq:operand-fold}
\end{equation}
We call this \emph{operand-fold} (\Cref{fig:fold-operand}): the scale multiplies the other input $A$, and the offset contributes a separate reduction of $A$ without reconstructing individual entries of $q$.

\input{figures/fig_fold_kv}

When each column $j$ of $q$ shares a scale $s_j$ and offset $z_j$ across the reduction index $k$, both terms factor out of the sum:
\begin{equation}
  \begin{aligned}
    C_{i,j} &= \sum_k A_{i,k}(s_jq_{k,j}+z_j) \\
      &= \underbrace{s_j\sum_k A_{i,k}q_{k,j}}_{\text{scaled partial result}}+
         \underbrace{z_j\sum_k A_{i,k}}_{\text{output correction}}.
  \end{aligned}
  \label{eq:acc-fold}
\end{equation}
We call this \emph{acc-fold} (\Cref{fig:fold-acc}): the scale multiplies the accumulated product, and the offset multiplies the corresponding row sum of $A$.
Both \emph{operand-fold} and \emph{acc-fold} preserve the result of applying $sq+z$ element-wise in exact arithmetic.

In attention, K and V arrive as $[L,D]$ tiles, with $L$ cached tokens and head dimension $D$.
Token-wise groups share a scale and offset along $D$ within each token, whereas channel-wise groups share them along $L$ within each channel.
Because $QK^{\!\top}$ reduces over $D$, token-wise K uses acc-fold and channel-wise K uses operand-fold into $Q$ (\Cref{fig:fold-kv-qk}).
Because $PV$ reduces over $L$, token-wise V uses operand-fold into $P$ and channel-wise V uses acc-fold (\Cref{fig:fold-kv-pv}).
\sys{} allows K and V to select their quantization and folding schemes independently according to the cache's quantization policy.
\Cref{fig:fold-pipeline} illustrates where scales and offsets are applied in the attention algorithm.
\sys{} retains online softmax~\cite{dao2023flashdecoding,dao2022flashattention,dao2024flashattention2} and changes only the QK and PV operand paths and their output corrections.

\input{figures/fig_fold_pipeline}

With $h_q$ query heads per KV head and group size $g$, affine folding reduces scalar scale-and-offset operations from one multiply-add per stored value to $2h_q/g$ (\Cref{sec:appendix-head-scaling}).

\subsection{Affine folding spans multiple quantization groups}
\label{sec:design-groups}

Under acc-fold, an unscaled partial product cannot cross a quantization-group boundary.
When a reduction spans multiple groups, the kernel applies each group's scale and offset to its partial product before adding the result to the running accumulator (\Cref{eq:acc-fold}).
We call this partial product \emph{group-local} because it sums only over the values that share one scale and offset.
Operand-fold instead scales the other matrix input before multiplication, so the accumulator can span multiple groups without rescaling; its offset correction still follows \Cref{eq:operand-fold}.

Both the Hopper and Blackwell kernels enforce the same ordering constraints on these corrections, although they assign the work to different warpgroups (\Cref{sec:design-pipelines}).
For operand-fold, the kernel can accumulate offset contributions for the same output separately and add their sum later.
QK corrections must precede softmax, while PV corrections must precede division by the accumulated softmax denominator.

\subsection{Constructing Tensor Core operands from packed integers}
\label{sec:design-unpack}

\sys{} constructs floating-point Tensor Core operands directly from packed integer bit patterns.%
\footnote{The Hopper kernel uses BF16 operands with \texttt{wgmma}.
The Blackwell kernel uses \texttt{tcgen05} with E4M3 inputs for $Q$ and $P$, E4M3 KV operands for INT4, and E2M1 KV operands for INT2 (using the \texttt{kind::f8f6f4} instruction mode~\cite{nvidia2026ptx}).}
Hopper combines each integer value with a fixed BF16 bit pattern, a technique known as mantissa injection~\cite{kim2022elephants,frantar2025marlin,du2026bitdecoding}.
Combining $q\in\{0,\ldots,15\}$ with \texttt{0x4300} via bitwise OR produces the exact BF16 value $y=q+128$.

On Blackwell, \sys{} can convert integer values into floating point more efficiently by \emph{directly placing each integer's bit pattern in the low bits of a floating-point operand and filling the remaining bits with zeros}.
For example, the bit patterns of the INT2 values $0,1,2,3$ represent $0,0.5,1,1.5$ in FP4 (E2M1), giving $y=q/2$.
INT4 similarly yields FP8 (E4M3) operands with $y=q\cdot2^{-9}$~\cite{lin2025marlinw4a8}.

\input{tables/operand-construction}
\input{figures/fig_dataflow}

These constructions produce operands of the form $y=\alpha q+\beta$ ($\alpha\ne0$) whose numerical values differ from the stored integers.
To account for $(\alpha,\beta)$ without extra per-element instructions, we fold both constants into each group's shared scale and offset.
Substituting $q=(y-\beta)/\alpha$ into $x=sq+z$ (\Cref{eq:dequant}) gives
\begin{equation}
  x=s'y+z',\qquad s'=s/\alpha,\qquad z'=z-\beta s'.
  \label{eq:operand-transform}
\end{equation}
Affine folding then applies $s'$ and $z'$ in place of $s$ and $z$, leaving the attention result unchanged in exact arithmetic (\Cref{sec:design-fold}).
$s'$ and $z'$ are precomputed in FP32 when the quantized KV cache is stored in HBM.
\Cref{tab:operand-construction} summarizes the $q\to y$ mappings, the $(\alpha,\beta)$ pairs used for compensation, and the ALU cost of converting each integer KV element into a Tensor Core operand.

\Cref{fig:dataflow-packing} shows how \sys{} packs INT2 K values in each 32-bit register to match these bit-level constructions.
On Hopper, each byte interleaves pairs of 2-bit values so that one bit shift and one masked merge (\texttt{LOP3}) produce two BF16 operands in a 32-bit register.
On Blackwell, each 32-bit register holds 16 two-bit values arranged as eight channel pairs; one bit shift and two masks unpack the register directly into two 32-bit registers holding 16 FP4 operands.
Both kernels use the Tensor Memory Accelerator (TMA) to copy packed tiles from global to shared memory (\Cref{fig:dataflow-threads}).
Hopper consumers build K operands in registers and collectively issue \texttt{wgmma}; Blackwell prepares a shared-memory operand tile that a separate issue thread supplies to Tensor Cores by issuing \texttt{tcgen05}~\cite{nvidia2026ptx}.
\Cref{sec:operand-construction} extends the INT2 K example to the packing and dataflow of INT2 V and INT4 KV operands on both GPUs.

\input{figures/fig_arch_timeline}

\subsection{Warp-specialized GPU pipelines}
\label{sec:design-pipelines}

We design separate warp-specialized pipelines for Hopper and Blackwell.
Hopper's \texttt{wgmma} accumulates into registers owned by its issuing warpgroup, whereas Blackwell's \texttt{tcgen05} accumulates into tensor memory (TMEM) accessible to other warpgroups in the CTA~\cite{nvidia2026ptx}.

On Hopper, one producer warpgroup issues TMA copies, and two consumer warpgroups process alternating even and odd KV tiles (\Cref{fig:arch-hopper}).
Each consumer builds KV operands and runs its own QK--softmax--PV chain, keeping online-softmax state and output accumulators in registers.
Each consumer applies scales and offset corrections according to \Cref{fig:fold-pipeline}: K corrections precede softmax, and V corrections update the output as described in \Cref{sec:design-groups}.
For channel-wise K, the producer warpgroup prepares the scaled $Q$ inputs.
At the end of the sequence, one consumer merges the two warpgroups' softmax states and partial outputs into the final CTA output.

Blackwell assigns copy and matrix-product issue, operand preparation, softmax, and output accumulation to separate warpgroups (\Cref{fig:arch-blackwell}).
The issue warpgroup starts TMA transfers and \texttt{tcgen05} products, while the preparation warpgroup constructs KV operands in shared memory.
The softmax warpgroup reads QK accumulators from TMEM, applies K offset corrections (and K scales for token-wise K) in FP32, computes the softmax weights $P$, and scales $P$ for token-wise V (quantizing folded $Q$ and $P$ tiles to E4M3 with a per-row scale).
The output warpgroup reads PV accumulators from TMEM, maintains the running output, and applies V offset corrections (and V scales for channel-wise V); for channel-wise K, it also prepares the scaled $Q$ inputs.
Shared-memory buffers pass operands between warpgroups, while TMEM passes matrix-product accumulators.

Multi-stage buffers in shared memory and TMEM allow producer warpgroups to work on later tiles while downstream warpgroups process earlier tiles.
We tune the buffer depth and producer lead distance empirically for each workload; for clarity, \Cref{fig:arch-blackwell} illustrates a one-step lead.

%% file: figures/fig_fold_mechanism.tex
\input{figures/fold-mechanism-common}
\def\fca{0.55}
\def\fcb{0.34}
\def\fcp{0.24}
\def\fcc{0.57}
\def\fcd{1.22}
\def\frh{0.38}
\def\dxs{3.56}
\tikzset{
  fdlab/.style={font=\scriptsize\sffamily, text=black!75, align=center, inner sep=1pt},
}
\providecommand{\fgrid}[6]{%
  \foreach \gi in {1,...,#4}{\ifnum\gi<#4
    \draw[mgrid] ({#1+\gi*#5},#2) -- ({#1+\gi*#5},{#2+#3*#6});\fi}%
  \foreach \gj in {1,...,#3}{\ifnum\gj<#3
    \draw[mgrid] (#1,{#2+\gj*#6}) -- ({#1+#4*#5},{#2+\gj*#6});\fi}%
  \draw[mframe] (#1,#2) rectangle ({#1+#4*#5},{#2+#3*#6});}
\providecommand{\ftint}[4]{%
  \fill[ptblue!26] (#1,#2) rectangle ++(#3,#4);}
\providecommand{\fcolvals}[4]{%
  \foreach \t [count=\i from 0] in {#4}{%
    \node[fcell] at ({#1+0.5*#3},{#2-(\i+0.5)*\frh}) {\t};}}
\providecommand{\frowvals}[4]{%
  \foreach \t [count=\i from 0] in {#4}{%
    \node[fcell] at ({#1+(\i+0.5)*#3},{#2+0.5*\frh}) {\t};}}
\providecommand{\foldbody}{%
  \fgrid{-1.91}{0}{3}{3}{\fca}{\frh}%
  \fgrid{0}{1.74}{3}{3}{\fcb}{\frh}%
  \fgrid{0}{0}{3}{3}{\fcb}{\frh}%
  \node[mname, anchor=south west] at (-1.91,1.22) {$A$};%
  \node[mname, anchor=west] at (1.13,2.24) {$q$};%
  \node[mname, anchor=south] at (0.51,1.22) {$C$};}
\providecommand{\foldaxes}{%
  \draw[axarrow] (-0.13,2.86) -- (-0.13,1.76);%
  \node[axname, anchor=east] at (-0.21,2.31) {$K$};}
\providecommand{\foldafull}{%
  \frowvals{-1.91}{0.76}{\fca}{{$A_{11}$},{$A_{12}$},{$A_{13}$}}%
  \frowvals{-1.91}{0.38}{\fca}{{$A_{21}$},{$A_{22}$},{$A_{23}$}}%
  \frowvals{-1.91}{0.00}{\fca}{{$A_{31}$},{$A_{32}$},{$A_{33}$}}}
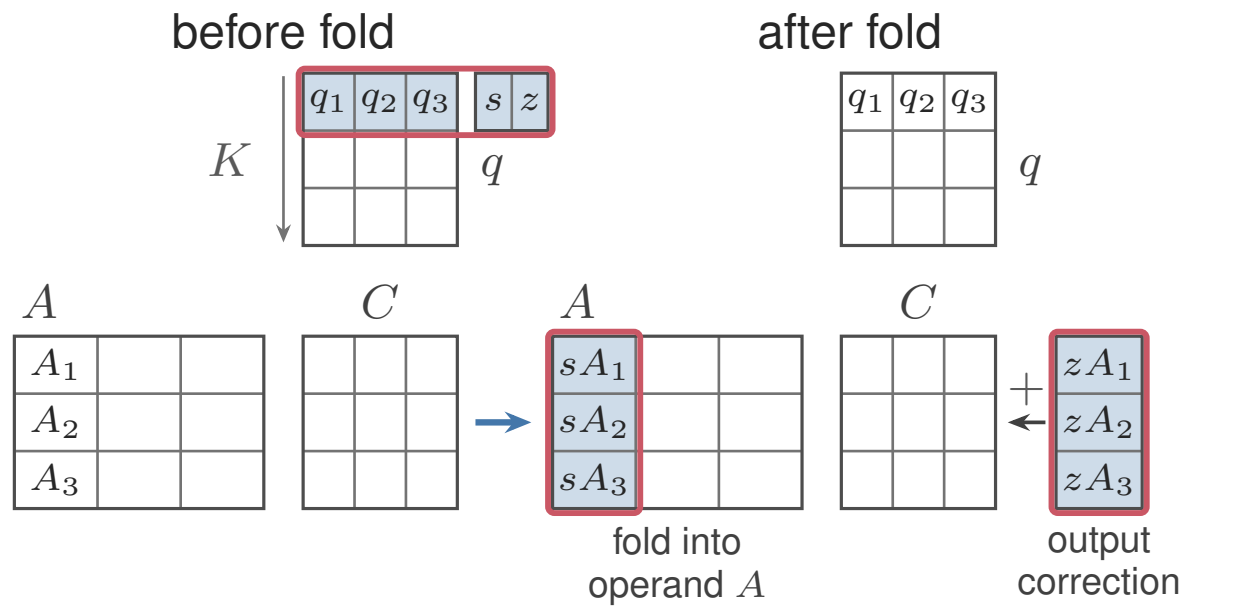
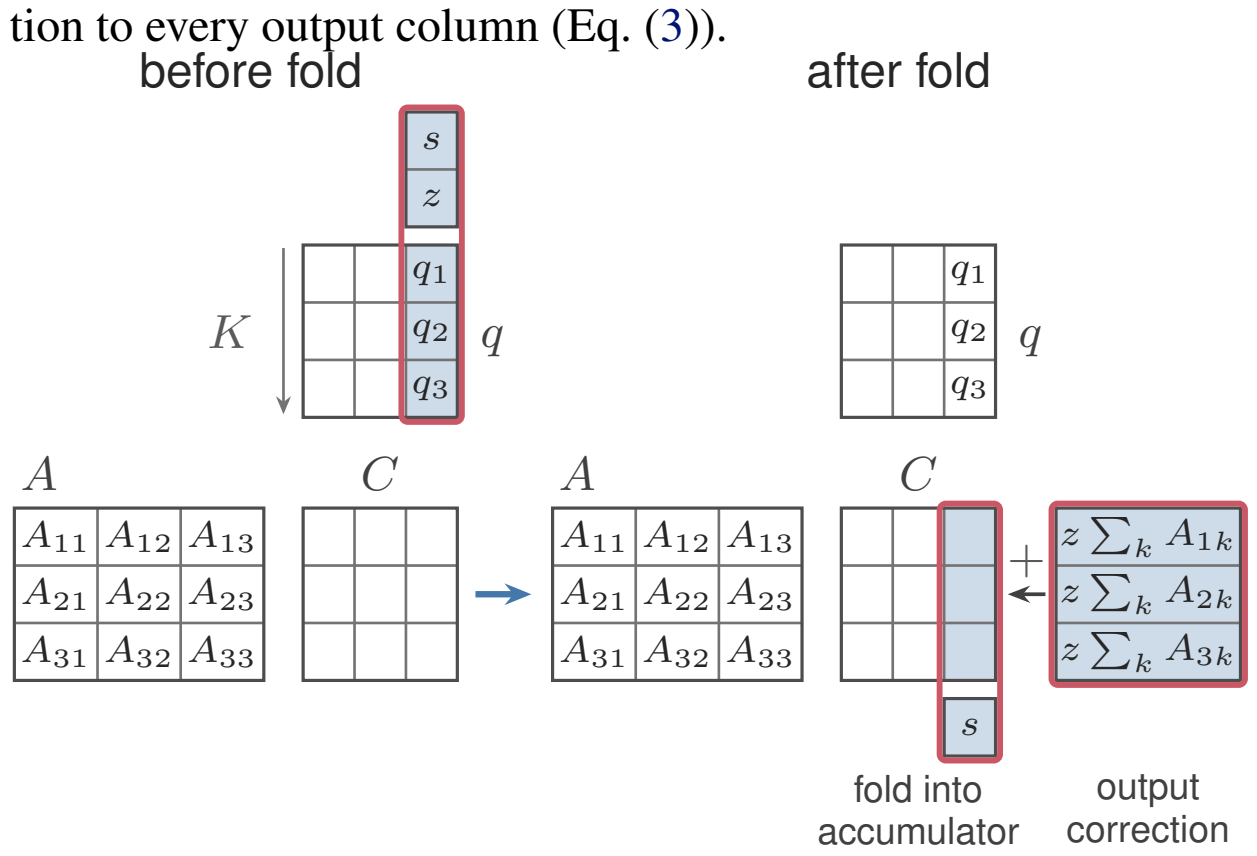
\begin{figure}[t]
\centering
\subfloat[Operand-fold scales a column of $A$ and adds the offset correction to every output column (\Cref{eq:operand-fold}).\label{fig:fold-operand}]{%
\begin{tikzpicture}[x=1cm, y=1cm, mname/.append style={inner sep=1pt},
                    caselab/.append style={inner sep=1pt}]
  \useasboundingbox (-1.94,3.30) rectangle (6.26,-0.64);
  \ftint{0}{2.50}{1.02}{\frh}
  \ftint{1.14}{2.50}{0.48}{\frh}
  \fgrid{1.14}{2.50}{1}{2}{\fcp}{\frh}
  \frowvals{1.14}{2.50}{\fcp}{{$s$},{$z$}}
  \foldbody
  \frowvals{0}{2.50}{\fcb}{{$q_1$},{$q_2$},{$q_3$}}
  \fcolvals{-1.91}{1.14}{\fca}{{$A_1$},{$A_2$},{$A_3$}}
  \foldaxes
  \draw[grpout] (-0.03,2.47) rectangle (1.65,2.91);
  \node[caselab, anchor=south] at (-0.13,3.00) {before fold};
  \begin{scope}[shift={(\dxs,0)}]
    \ftint{-1.91}{0}{\fca}{1.14}
    \ftint{1.42}{0}{\fcc}{1.14}
    \fgrid{1.42}{0}{3}{1}{\fcc}{\frh}
    \fcolvals{1.42}{1.14}{\fcc}{{$zA_1$},{$zA_2$},{$zA_3$}}
    \foldbody
    \frowvals{0}{2.50}{\fcb}{{$q_1$},{$q_2$},{$q_3$}}
    \fcolvals{-1.91}{1.14}{\fca}{{$sA_1$},{$sA_2$},{$sA_3$}}
    \draw[flow] (1.35,0.57) -- (1.10,0.57);
    \node[axname, anchor=south, inner sep=0.5pt] at (1.225,0.66) {$+$};
    \draw[grpout] (-1.94,-0.03) rectangle (-1.33,1.17);
    \draw[grpout] (1.39,-0.03) rectangle (2.02,1.17);
    \node[fdlab, anchor=north] at (-1.085,-0.08) {fold into\\operand $A$};
    \node[fdlab, anchor=north] at (1.705,-0.08) {output\\correction};
    \node[caselab, anchor=south] at (0.055,3.00) {after fold};
  \end{scope}
  \draw[foldbig] (1.14,0.57) -- (1.51,0.57);
\end{tikzpicture}}
\\[-2pt]
\subfloat[Acc-fold scales a partial output column and adds the offset-weighted row sums of $A$ (\Cref{eq:acc-fold}).\label{fig:fold-acc}]{%
\begin{tikzpicture}[x=1cm, y=1cm, mname/.append style={inner sep=1pt},
                    caselab/.append style={inner sep=1pt}]
  \useasboundingbox (-1.94,4.16) rectangle (6.26,-1.14);
  \ftint{0.68}{1.74}{\fcb}{1.14}
  \ftint{0.68}{3.00}{\fcb}{0.76}
  \fgrid{0.68}{3.00}{2}{1}{\fcb}{\frh}
  \fcolvals{0.68}{3.76}{\fcb}{{$s$},{$z$}}
  \foldbody
  \fcolvals{0.68}{2.88}{\fcb}{{$q_1$},{$q_2$},{$q_3$}}
  \foldafull
  \foldaxes
  \draw[grpout] (0.65,1.71) rectangle (1.05,3.79);
  \node[caselab, anchor=south] at (-0.345,3.88) {before fold};
  \begin{scope}[shift={(\dxs,0)}]
    \ftint{0.68}{0}{\fcb}{1.14}
    \ftint{0.68}{-0.50}{\fcb}{\frh}
    \fgrid{0.68}{-0.50}{1}{1}{\fcb}{\frh}
    \fcolvals{0.68}{-0.12}{\fcb}{{$s$}}
    \ftint{1.42}{0}{\fcd}{1.14}
    \fgrid{1.42}{0}{3}{1}{\fcd}{\frh}
    \fcolvals{1.42}{1.14}{\fcd}{{$z\textstyle\sum_k A_{1k}$},{$z\textstyle\sum_k A_{2k}$},{$z\textstyle\sum_k A_{3k}$}}
    \foldbody
    \fcolvals{0.68}{2.88}{\fcb}{{$q_1$},{$q_2$},{$q_3$}}
    \foldafull
    \draw[flow] (1.35,0.57) -- (1.10,0.57);
    \node[axname, anchor=south, inner sep=0.5pt] at (1.225,0.66) {$+$};
    \draw[grpout] (0.65,-0.53) rectangle (1.05,1.17);
    \draw[grpout] (1.39,-0.03) rectangle (2.67,1.17);
    \node[fdlab, anchor=north] at (0.51,-0.58) {fold into\\accumulator};
    \node[fdlab, anchor=north] at (2.03,-0.58) {output\\correction};
    \node[caselab, anchor=south] at (0.38,3.88) {after fold};
  \end{scope}
  \draw[foldbig] (1.14,0.57) -- (1.51,0.57);
\end{tikzpicture}}
\caption{Affine folding of $C=A(sq+z)$, where $K$ is the reduction axis.}
\label{fig:fold-mechanism}
\end{figure}

%% file: figures/fold-mechanism-common.tex
\tikzset{
  mgrid/.style={draw=black!55, line width=0.5pt},
  mframe/.style={draw=black!70, line width=0.6pt},
  mname/.style={font=\footnotesize\sffamily, text=black!75},
  fcell/.style={font=\scriptsize, text=black!85, inner sep=0pt},
  axname/.style={font=\footnotesize\sffamily, text=black!65},
  axarrow/.style={-{Stealth[length=3.4pt, width=2.8pt]}, draw=black!55, line width=0.5pt},
  szarrow/.style={{Stealth[length=3.2pt, width=2.6pt]}-{Stealth[length=3.2pt, width=2.6pt]}, draw=ptred!85!black, line width=0.7pt},
  szlab/.style={font=\footnotesize\sffamily, text=ptred!70!black, fill=white, inner sep=0.6pt},
  caselab/.style={font=\footnotesize\sffamily, text=black!85, align=center},
  mode/.style={font=\footnotesize\sffamily\bfseries, text=ptblue!80!black, align=center},
  qbox/.style={draw=black!70, line width=0.5pt, rounded corners=2pt, minimum height=0.54cm, minimum width=0.72cm, inner xsep=3pt, align=center, font=\footnotesize\sffamily},
  quant/.style={qbox, double, double distance=1pt, fill=ptcyan!20},
  prod/.style={qbox, fill=white},
  res/.style={qbox, fill=ptgreen!16},
  flow/.style={-{Stealth[length=4.2pt, width=3.2pt]}, draw=black!75, line width=0.7pt},
  foldline/.style={-{Stealth[length=4pt, width=3pt]}, draw=ptblue, line width=0.8pt, dashed},
  grpout/.style={draw=ptred!85!black, line width=1.1pt, rounded corners=1.2pt},
  foldbig/.style={-{Stealth[length=5pt, width=4pt]}, draw=ptblue, line width=1.2pt},
  foldlab/.style={font=\footnotesize\sffamily, text=ptblue!80!black, fill=white, inner sep=1pt, align=center},
}

\providecommand{\foldmgridd}[5]{%
  \foreach \gi in {1,...,#4}{\ifnum\gi<#4
    \draw[mgrid] (#1+\gi*#5,#2) -- (#1+\gi*#5,#2+#3*#5);\fi}
  \foreach \gj in {1,...,#3}{\ifnum\gj<#3
    \draw[mgrid] (#1,#2+\gj*#5) -- (#1+#4*#5,#2+\gj*#5);\fi}
  \draw[mframe] (#1,#2) rectangle ++(#4*#5,#3*#5);}

%% file: figures/fig_fold_kv.tex
\input{figures/fold-mechanism-common}
\newcommand{\foldkvpanel}[2]{%
\begin{tikzpicture}[x=1cm, y=1cm]
  \def\t{0.38}
  \def\my{0.36}
  \useasboundingbox (-0.20,1.80) rectangle (8.04,-0.44);
  \foreach \r [count=\i from 0] in {1,...,3}{\pgfmathsetmacro{\tt}{14+24*\i}
    \fill[ptblue!\tt] (0.46,\my+3*\t-\i*\t-\t) rectangle ++(3*\t,\t);}
  \foldmgridd{0.46}{\my}{3}{3}{\t}
  \node[axname, anchor=east]  at (0.40,\my+1.5*\t) {$L$};
  \node[axname, anchor=south] at (0.46+1.5*\t,\my+3*\t+0.02) {$D$};
  \draw[szarrow] (0.50,\my+1.5*\t) -- (0.46+3*\t-0.04,\my+1.5*\t);
  \node[caselab, anchor=north] at (0.46+1.5*\t,\my-0.08) {token-wise\\$s,z$ span $D$};
  \draw[foldline] (1.72,\my+1.5*\t) -- (1.96,\my+1.5*\t);
  \node[mode, anchor=west] at (2.02,\my+1.5*\t) {#1};
  \foreach \q [count=\j from 0] in {1,...,3}{\pgfmathsetmacro{\tt}{14+24*\j}
    \fill[ptblue!\tt] (4.30+\j*\t,\my) rectangle ++(\t,3*\t);}
  \foldmgridd{4.30}{\my}{3}{3}{\t}
  \node[axname, anchor=south] at (4.30+1.5*\t,\my+3*\t+0.02) {$D$};
  \draw[szarrow] (4.30+1.5*\t,\my+0.04) -- (4.30+1.5*\t,\my+3*\t-0.04);
  \node[caselab, anchor=north] at (4.30+1.5*\t,\my-0.08) {channel-wise\\$s,z$ span $L$};
  \draw[foldline] (5.56,\my+1.5*\t) -- (5.80,\my+1.5*\t);
  \node[mode, anchor=west] at (5.86,\my+1.5*\t) {#2};
\end{tikzpicture}}
\begin{figure}[t]
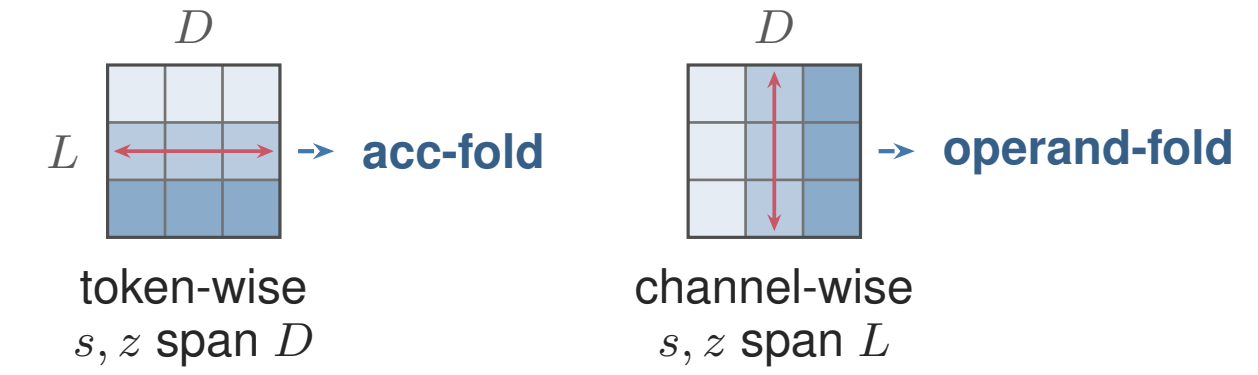
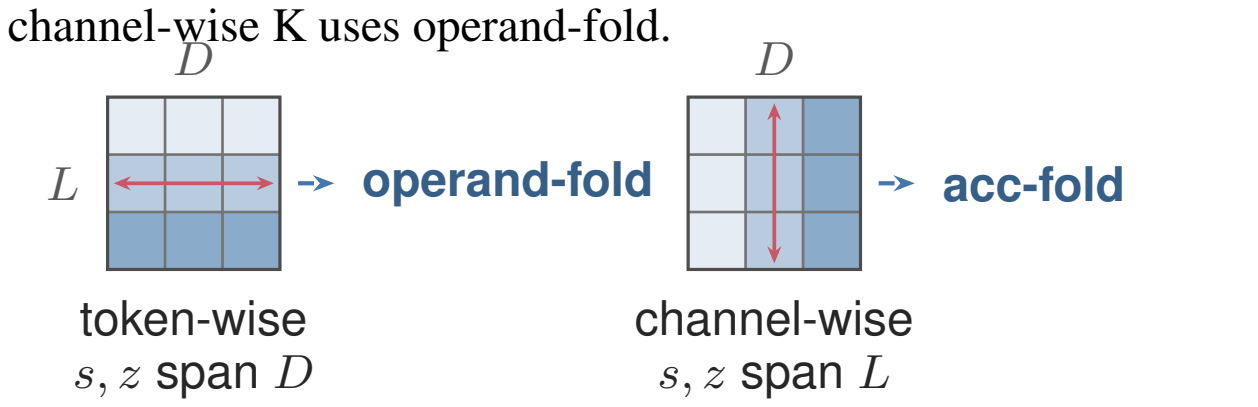

\centering
\subfloat[$QK^{\!\top}$ reduces over $D$, so token-wise K uses acc-fold and channel-wise K uses operand-fold.\label{fig:fold-kv-qk}]{\foldkvpanel{acc-fold}{operand-fold}}\\[-2pt]
\subfloat[$PV$ reduces over $L$, so token-wise V uses operand-fold and channel-wise V uses acc-fold.\label{fig:fold-kv-pv}]{\foldkvpanel{operand-fold}{acc-fold}}
\caption{Affine folding modes for token-wise and channel-wise K and V tiles.}
\label{fig:fold-kv}
\end{figure}

%% file: figures/fig_fold_pipeline.tex
\input{figures/fold-mechanism-common}
\begin{figure}[t]
\centering
\begin{tikzpicture}[x=1cm, y=1cm, qbox/.append style={minimum height=0.46cm}]
  \node[prod]  (Q)  at (0.95, 2.70) {$Q$};
  \node[prod]  (qk) at (0.95, 2.04) {QK};
  \node[res]   (S)  at (0.95, 1.38) {$S$};
  \node[prod]  (P)  at (0.95, 0.72) {$P$};
  \node[prod]  (pv) at (0.95, 0.06) {PV};
  \node[res]   (O)  at (0.95,-0.60) {$O$};
  \node[quant] (Kq) at (-4.00, 2.04) {$K^{\!\top}$ quantized};
  \node[quant] (Vq) at (-4.00, 0.06) {$V$ quantized};
  \node[prod]  (Ky) at (-1.45, 2.04) {K operand};
  \node[prod]  (Vy) at (-1.45, 0.06) {V operand};
  \draw[flow] (Q) -- (qk);
  \draw[flow] (qk) -- (S);
  \draw[flow] (S) -- node[axname, right, xshift=0.42cm, inner sep=1pt] {softmax} (P);
  \draw[flow] (P) -- (pv);
  \draw[flow] (pv) -- (O);
  \node[axname, anchor=west] at (1.62,1.38) {K corr.};
  \draw[flow] (1.55,1.38) -- (S.east);
  \node[axname, anchor=west] at (1.62,-0.60) {V corr.};
  \draw[flow] (1.55,-0.60) -- (O.east);
  \draw[flow] (Kq) -- (Ky);
  \draw[flow] (Ky) -- (qk);
  \draw[flow] (Vq) -- (Vy);
  \draw[flow] (Vy) -- (pv);
  \foreach \src/\ysrc/\ydst/\lab in {Kq/2.31/2.70/{channel-wise K\\operand-fold}, Kq/1.77/1.38/{token-wise K\\acc-fold}, Vq/0.33/0.72/{token-wise V\\operand-fold}, Vq/-0.21/-0.60/{channel-wise V\\acc-fold}}{
    \draw[foldline] (-4.00,\ysrc) -- (-4.00,\ydst) -- node[foldlab, pos=0.52] {\lab} (0.55,\ydst);}
\end{tikzpicture}
\caption{Affine folding in attention.
Dashed arrows show where each K and V quantization scheme applies its scales.}
\label{fig:fold-pipeline}
\end{figure}
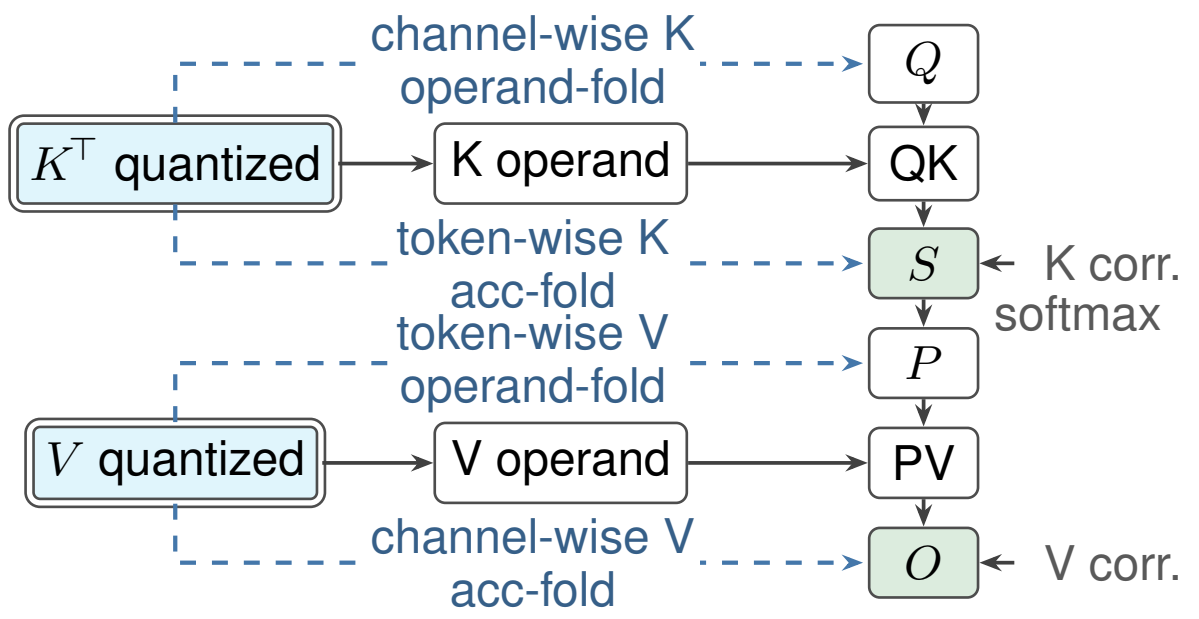

%% file: tables/operand-construction.tex
\begin{table}[t]
  \centering
  \scriptsize
  \begin{tabular*}{\columnwidth}{@{\extracolsep{\fill}}lcccc@{}}
    \toprule
    Mapping & $q\to y$ & $y$ & $(\alpha,\beta)$ & ALU/elem. \\
    \midrule
    \multicolumn{5}{@{}l}{\emph{Hopper}} \\
    \cmidrule(lr){1-5}
    INT4$\to$BF16 & $\texttt{0x4300}\,|\,q$ & $q+128$ & $(1,128)$ & 1.00 \\
    INT2$\to$BF16 & $\texttt{0x4300}\,|\,q$ & $q+128$ & $(1,128)$ & 1.00 \\
    \multicolumn{5}{@{}l}{\emph{Blackwell}} \\
    \cmidrule(lr){1-5}
    INT4$\to$E4M3 & $q$ & $q\cdot2^{-9}$ & $(2^{-9},0)$ & 0.38 \\
    INT2$\to$E2M1 & $q$ & $q/2$ & $(1/2,0)$ & 0.19 \\
    \bottomrule
  \end{tabular*}
  \caption{Tensor Core operand mappings and their static ALU costs per stored value.
  The $q{\to}y$ column gives the bit-level construction of $y$ from $q$.}
  \label{tab:operand-construction}
\end{table}

%% file: figures/fig_dataflow.tex
\begin{figure*}[t]
\centering
\subfloat[Packed register layout and bit-level operand construction.\label{fig:dataflow-packing}]{%
\begin{minipage}[t]{0.49\textwidth}
\centering
\includegraphics[width=\linewidth]{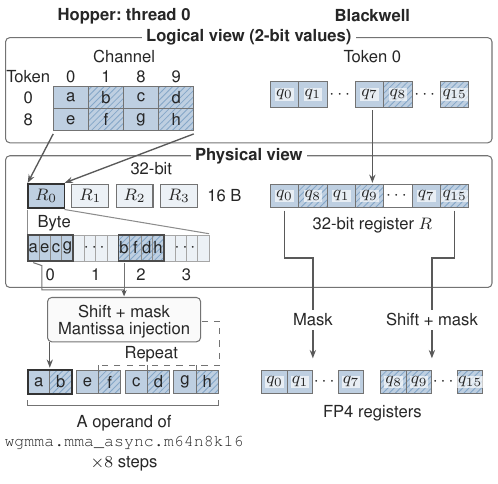}
\end{minipage}}%
\hfill
\subfloat[K operand flow from global memory to the Tensor Core.
Tile shapes are logical (tokens $\times$ channels).\label{fig:dataflow-threads}]{%
\begin{minipage}[t]{0.49\textwidth}
\centering
\includegraphics[width=\linewidth]{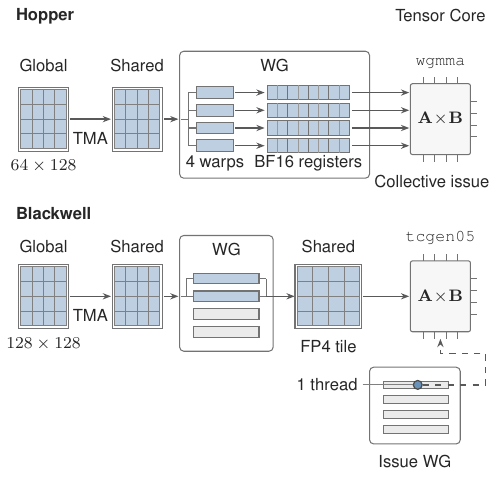}
\end{minipage}}
\caption{Register packing and global-to-Tensor-Core dataflow for INT2 K in \sys{}.}
\label{fig:dataflow}
\end{figure*}

%% file: figures/fig_arch_timeline.tex
\begin{figure*}[t]
\centering
\subfloat[H100: \texttt{WG0} issues the tile copies using TMA, and two consumer warpgroups alternate even and odd tiles.\label{fig:arch-hopper}]{%
  \includegraphics[width=\textwidth]{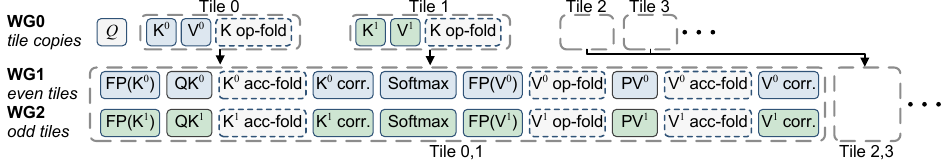}}

\subfloat[B200: \texttt{WG0} issues the tile copies and \texttt{tcgen05} matrix instructions, while \texttt{WG1}--\texttt{WG3} prepare KV operands, apply affine folding, and compute softmax on CUDA Cores.\label{fig:arch-blackwell}]{%
  \includegraphics[width=\textwidth]{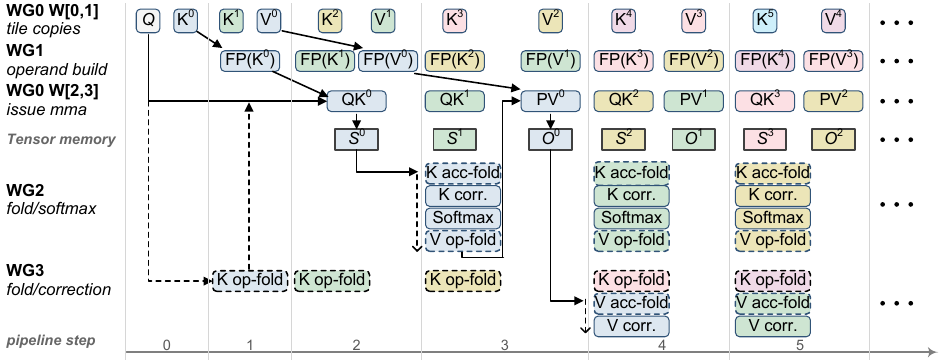}}
\caption{\sys{}'s Hopper and Blackwell pipelines.
$\mathrm{FP}(X)$ constructs $X$'s floating-point Tensor Core operand from packed integers (\Cref{sec:design-unpack}).
\texttt{corr.}\ denotes the output correction, and \texttt{WG}$x$ \texttt{W[}$y$\texttt{]} denotes warpgroup $x$, warp $y$.
Dashed boxes execute only for the corresponding K or V quantization axis.}
\label{fig:arch-timeline}
\end{figure*}

%% file: src/4-serving.tex
\section{Serving Engine Integration}
\label{sec:serving}

We integrate \sys{} into SGLang as a backend for paged decode attention~\cite{zheng2024sglang}.
\sys{} incorporates common optimization techniques, including split-KV, which divides long contexts across CTAs to increase parallelism.
To better understand how \sys{} benefits existing serving systems with quantized KV caches, we integrate \sys{} with OSCAR~\cite{zhou2026oscar} as a drop-in replacement for its INT2 kernel and keep OSCAR's quantization policy, rotations, and BF16 windows (\Cref{sec:eval-serving-throughput}) unchanged.

%% file: src/5-evaluation.tex
\section{Evaluation}
\label{sec:eval}

We implement \sys{} in CuTe DSL (CUTLASS~4.8, CUDA~13.0) and evaluate it on NVIDIA H100 (80~GB HBM3) and B200 (180~GB HBM3e) GPUs with $H_Q=32$, $H_{KV}=8$, $D=128$, page size 64, and token-wise K and V quantization groups of $g=128$ values with FP32 scales and offsets.
Token-wise K and V exercise both acc-fold on QK and operand-fold on PV; \Cref{sec:appendix-axes} evaluates all four combinations of token-wise and channel-wise K and V, where \sys{} achieves $1.75$--$4.00\times$ speedups over FlashInfer BF16 on H100 and $1.50$--$2.74\times$ on B200.
We evaluate batch sizes 1 and 8 at context lengths of 8K, 32K, 128K, and 512K, and batch size 64 at 8K, 16K, 32K, 64K, and 128K.
Each measurement takes the median of five timed iterations (excluding an L2 flush before each call) after five warmups, and includes all decode kernels and split reduction.
Plotted speedups normalize each kernel's latency to FlashInfer BF16, whereas speedups quoted in the text compare \sys{} directly against each named baseline within the same benchmark series (using five-run medians for B200 INT4 vendor baselines and three-run medians for BitDecoding and SAW).

\subsection{Kernel performance}
\label{sec:eval-performance}

\input{figures/speedup-subbyte}
\input{figures/speedup-wide}

\Cref{fig:speedup-subbyte,fig:speedup-wide} compare \sys{} against sub-byte and wider-format baselines.
On H100, \sys{} INT2 achieves a $3.36\times$ geometric-mean speedup over FlashInfer BF16 and $2.03\times$ over TensorRT-LLM FP8, with maxima of $4.15\times$ and $2.44\times$.
On B200, the geometric-mean speedups are $2.67\times$ over BF16, $1.48\times$ over TensorRT-LLM FP8, and $1.16\times$ over TensorRT-LLM NVFP4, with maxima of $3.25\times$, $1.64\times$, and $1.26\times$.
These vendor kernels are invoked through FlashInfer, and \sys{} INT2 outperforms them across all tested batch sizes and context lengths.
Against OSCAR INT2 and BitDecoding INT2, \sys{} INT2 achieves geometric-mean speedups of $2.31\times$ and $1.77\times$ on H100, and $3.44\times$ and $2.78\times$ on B200.

On H100, \sys{} INT4 achieves geometric-mean speedups of $2.83\times$ over FlashInfer BF16, $1.71\times$ over TensorRT-LLM FP8, $1.32\times$ over BitDecoding INT4, and $4.19\times$ over SAW INT4.
On B200, \sys{} INT4 achieves geometric-mean speedups of $2.30\times$ over FlashInfer BF16, $1.28\times$ over TensorRT-LLM FP8, $1.94\times$ over BitDecoding INT4, and $6.93\times$ over SAW INT4.
It matches native NVFP4 overall ($1.00\times$, ranging from $0.88\times$ to $1.20\times$; faster at 32K--128K for batch 64), while returning BF16 output instead of NVFP4's FP8 output.
\Cref{sec:appendix-head-scaling} further evaluates how \sys{} INT2 speedup scales across $H_{KV}\in[1,32]$.

\input{figures/fold-ab-grid}

\subsection{Kernel accuracy}
\label{sec:eval-accuracy}

\input{figures/accuracy-arms}

\Cref{fig:accuracy-arms} compares the $L_2$ output error of BF16 decode (FlashInfer), \sys{} INT4, \sys{} INT2, and OSCAR INT2.
To stress-test quantization, our synthetic input amplifies eight K channels, ramps K's per-token scale along the sequence, and rescales K so that pre-softmax attention logits have unit standard deviation.
For INT2, \sys{} and OSCAR decode the same quantized cache written by OSCAR's clipped quantizer without BF16 windows.
Both INT2 kernels yield $L_2$ errors of $0.52$--$0.61$, compared with $0.16$--$0.22$ for \sys{} INT4 and $(2.13\text{--}2.25)\times10^{-3}$ for BF16 decode.
Under unclipped min--max INT2 quantization, the $L_2$ error rises to $0.86$--$1.03$.
Compared directly against the dequantized INT2 cache, \sys{}'s kernel arithmetic error is only $(1.68\text{--}1.76)\times10^{-3}$ on both H100 and B200, matching BF16 decode and confirming that the $0.52$--$0.61$ error stems from 2-bit quantization rather than affine folding.

\subsection{Kernel ablation}
\label{sec:eval-ablation}

On INT2 caches with equal split-KV counts within each pair, \Cref{fig:fold-ab-grid} ablates affine folding against per-value reconstruction on H100 and B200, and FP4 (E2M1) against FP8 (E4M3) KV Tensor Core operands on B200.
On H100, \sys{}'s speedup over external baselines combines its warp-specialized kernel design with affine folding, which adds a $1.07$--$1.15\times$ speedup over per-value reconstruction (\Cref{fig:fold-ab-hopper}) as expected when H100 reconstruction remains within the hardware issue peak (\Cref{fig:ops-per-byte}).
On B200, the reconstruction baseline follows the standard approach of dequantizing INT2 values into FP8 (E4M3) Tensor Core operands before each matrix product; because this per-value reconstruction exceeds B200's issue peak, affine folding achieves a $7.36$--$8.86\times$ speedup (\Cref{fig:fold-ab-b200-folding}).
Switching the folded B200 INT2 kernel from FP8 (E4M3) to FP4 (E2M1) KV operands and their matching packing layout provides an additional $1.17$--$1.21\times$ speedup (\Cref{fig:fold-ab-b200-feed}).

\subsection{Serving throughput}
\label{sec:eval-serving-throughput}

\input{figures/serving-pareto}

We serve Qwen3-8B with 100K input tokens, 1K output tokens, and 99\% or 95\% prefix-cache hits (one cached prefix per request), sharing a BF16 prefill backend across configurations.
We evaluate two INT2 configurations to show both realistic accuracy-preserving integration and pure 2-bit kernel speed: \emph{OSCAR INT2 + Ours} replaces OSCAR's INT2 decode kernel with \sys{} while keeping OSCAR's rotations and BF16 windows (128 initial and 512 recent tokens, $<0.7\%$ of the 100K context), whereas \emph{Ours INT2} uses \sys{}'s token-wise INT2 K and V cache without recent windows or rotations.
We vary concurrency in powers of two up to 16 on H100 and 32 (plus 48) on B200 to measure per-user (median request output tokens/s) and aggregate (total output tokens/s) throughput (\Cref{fig:serving-pareto}).

At a 99\% prefix-cache hit rate, \emph{OSCAR INT2 + Ours} reaches a peak aggregate throughput of 487 tokens/s on H100 and 868 tokens/s on B200, compared with 269 and 333 tokens/s for OSCAR ($81\%$ and $161\%$ higher).
\emph{Ours INT2} reaches 506 tokens/s on H100 and 897 tokens/s on B200, improving over OSCAR by $88\%$ and $170\%$; this also exceeds FlashInfer BF16's peaks of 115 tokens/s on H100 (at concurrency 2) and 355 tokens/s on B200 (at concurrency 8) by $341\%$ and $153\%$.
At 95\% prefix hits, \emph{OSCAR INT2 + Ours} and \emph{Ours INT2} reach 337 and 346 tokens/s versus OSCAR's 214 tokens/s on H100, and 738 and 755 tokens/s versus 297 tokens/s on B200.
Across cache formats, peak throughput reflects both decoder speed and the number of concurrent requests that fit in memory.

\subsection{Serving accuracy}
\label{sec:eval-serving-accuracy}

\input{tables/serving-accuracy}

\Cref{tab:serving-accuracy} compares FlashInfer BF16, OSCAR INT2, and \emph{OSCAR INT2 + Ours} on GPQA-Diamond~\cite{rein2023gpqa} and GSM8K on H100, reporting the mean and sample standard deviation across five runs per configuration.
Every question-paired 95\% confidence interval for the accuracy difference between configurations spans zero, so five runs show no measurable accuracy gain or loss (though five runs do not establish formal equivalence).
Median prompt lengths are 243 tokens for GPQA-Diamond and 91 tokens for GSM8K; because \emph{OSCAR INT2 + Ours} decodes the identical quantized cache as OSCAR with $<1.8\times10^{-3}$ arithmetic error (\Cref{sec:eval-accuracy}), it preserves OSCAR's cache accuracy while accelerating decode.

%% file: figures/speedup-subbyte.tex
\begin{figure}[t]
  \centering
  \captionsetup[subfloat]{farskip=2pt}
  \includegraphics[width=\columnwidth]{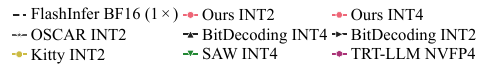}
  \par
  \subfloat[H100, batch 1.\label{fig:speedup-subbyte-h100-b1}]{%
    \includegraphics{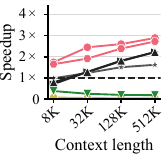}}
  \hfill
  \subfloat[H100, batch 8.\label{fig:speedup-subbyte-h100-b8}]{%
    \includegraphics{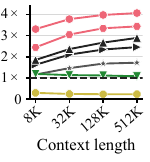}}
  \hfill
  \subfloat[H100, batch 64.\label{fig:speedup-subbyte-h100-b64}]{%
    \includegraphics{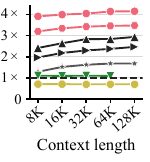}}
  \par\vspace{3pt}
  \subfloat[B200, batch 1.\label{fig:speedup-subbyte-b200-b1}]{%
    \includegraphics{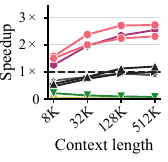}}
  \hfill
  \subfloat[B200, batch 8.\label{fig:speedup-subbyte-b200-b8}]{%
    \includegraphics{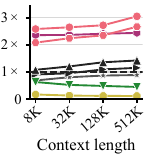}}
  \hfill
  \subfloat[B200, batch 64.\label{fig:speedup-subbyte-b200-b64}]{%
    \includegraphics{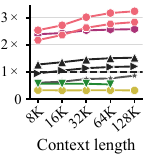}}
  \caption{Kernel speedups of \sys{} and 2- and 4-bit baselines over FlashInfer BF16.}
  \label{fig:speedup-subbyte}
\end{figure}

%% file: figures/speedup-wide.tex
\begin{figure}[t]
  \centering
  \captionsetup[subfloat]{farskip=2pt}
  \includegraphics[width=\columnwidth]{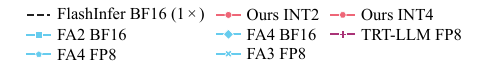}
  \par
  \subfloat[H100, batch 1.\label{fig:speedup-wide-h100-b1}]{%
    \includegraphics{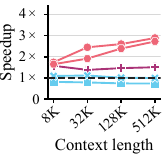}}
  \hfill
  \subfloat[H100, batch 8.\label{fig:speedup-wide-h100-b8}]{%
    \includegraphics{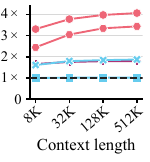}}
  \hfill
  \subfloat[H100, batch 64.\label{fig:speedup-wide-h100-b64}]{%
    \includegraphics{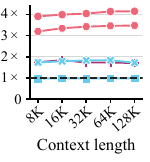}}
  \par\vspace{3pt}
  \subfloat[B200, batch 1.\label{fig:speedup-wide-b200-b1}]{%
    \includegraphics{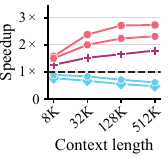}}
  \hfill
  \subfloat[B200, batch 8.\label{fig:speedup-wide-b200-b8}]{%
    \includegraphics{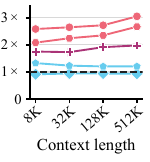}}
  \hfill
  \subfloat[B200, batch 64.\label{fig:speedup-wide-b200-b64}]{%
    \includegraphics{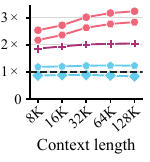}}
  \caption{Kernel speedups of \sys{} and 8- and 16-bit baselines over FlashInfer BF16.
  The dashed line marks parity.}
  \label{fig:speedup-wide}
\end{figure}

%% file: figures/fold-ab-grid.tex
\begin{figure}[t]
  \centering
  \captionsetup[subfloat]{farskip=2pt,justification=raggedright,singlelinecheck=false}
  \includegraphics[width=\columnwidth]{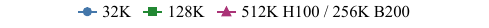}
  \par
  \begin{minipage}[t]{0.345\columnwidth}
  \centering
  \subfloat[H100: \mbox{reconstruction} versus folding.\label{fig:fold-ab-hopper}]{%
    \includegraphics[width=\linewidth]{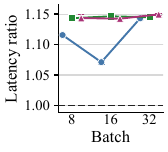}}
  \end{minipage}\hfill
  \begin{minipage}[t]{0.307\columnwidth}
  \centering
  \subfloat[B200: E4M3 \mbox{reconstruction} versus folding.\label{fig:fold-ab-b200-folding}\label{fig:fold-ab-grid-b200}]{%
    \includegraphics[width=\linewidth]{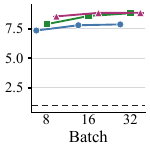}}
  \end{minipage}\hfill
  \begin{minipage}[t]{0.307\columnwidth}
  \centering
  \subfloat[B200: folded E4M3 versus E2M1.\label{fig:fold-ab-b200-feed}]{%
    \includegraphics[width=\linewidth]{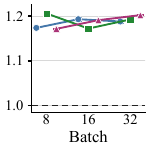}}
  \end{minipage}
  \caption{\sys{} INT2 kernel latency ratios between ablation variants at equal split-KV counts (first variant's latency divided by the second's).
  Panels (b) and (c) decode an INT2 cache into E4M3 or E2M1 Tensor Core operands on B200; error bars show paired 95\% confidence intervals across ten runs.}
  \label{fig:fold-ab-grid}
\end{figure}

%% file: figures/accuracy-arms.tex
\begin{figure}[t]
  \centering
  \includegraphics[width=\columnwidth]{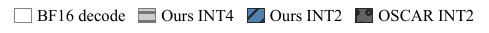}
  \par
  \subfloat[H100.\label{fig:accuracy-arms-h100}]{%
    \includegraphics[width=0.49\columnwidth]{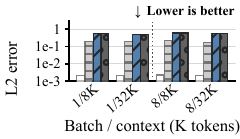}}
  \hfill
  \subfloat[B200.\label{fig:accuracy-arms-b200}]{%
    \includegraphics[width=0.49\columnwidth]{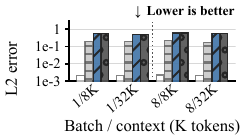}}
  \caption{$L_2$ output error of BF16 decode (FlashInfer), \sys{} INT4, \sys{} INT2, and OSCAR INT2 on the synthetic input (logarithmic scale).}
  \label{fig:accuracy-arms}
\end{figure}

%% file: figures/serving-pareto.tex
\begin{figure}[t]
  \centering
  \captionsetup[subfloat]{farskip=2pt}
  \includegraphics[width=\columnwidth]{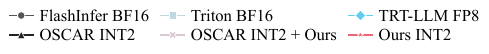}
  \par
  \subfloat[H100, 99\% prefix-cache hit.\label{fig:serving-pareto-h100-99}]{%
    \includegraphics{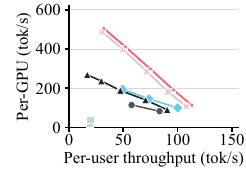}}
  \hfill
  \subfloat[H100, 95\% prefix-cache hit.\label{fig:serving-pareto-h100-95}]{%
    \includegraphics{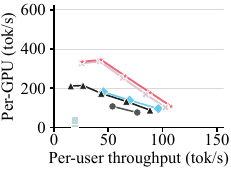}}
  \par\vspace{3pt}
  \subfloat[B200, 99\% prefix-cache hit.\label{fig:serving-pareto-b200-99}]{%
    \includegraphics{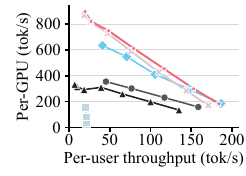}}
  \hfill
  \subfloat[B200, 95\% prefix-cache hit.\label{fig:serving-pareto-b200-95}]{%
    \includegraphics{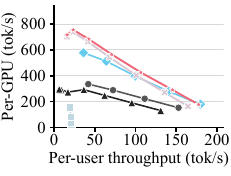}}
  \caption{Per-user and aggregate Qwen3-8B serving throughput on a single GPU with 100K input and 1K output tokens.}
  \label{fig:serving-pareto}
\end{figure}

%% file: tables/serving-accuracy.tex
\begin{table}[t]
  \centering
  \caption{Qwen3-8B serving accuracy on H100 (mean $\pm$ sample standard deviation across five runs).}
  \label{tab:serving-accuracy}
  \footnotesize
  \setlength{\tabcolsep}{4pt}
  \begin{tabular}{@{}lrr@{}}
    \toprule
    Configuration & GPQA-Diamond (\%) & GSM8K (\%) \\
    \midrule
    FlashInfer BF16 & $56.36 \pm 1.84$ & $95.71 \pm 0.11$ \\
    OSCAR INT2 & $54.65 \pm 2.18$ & $95.74 \pm 0.38$ \\
    OSCAR INT2 + Ours & $55.45 \pm 1.49$ & $95.92 \pm 0.15$ \\
    \bottomrule
  \end{tabular}
\end{table}

%% file: src/6-related.tex
\section{Related Work}
\label{sec:related}

\subsection{KV-cache quantization algorithms}

KIVI~\cite{liu2024kivi}, KVQuant~\cite{hooper2024kvquant}, and Kitty~\cite{xia2025kitty} establish low-bit KV formats and motivate channel-wise grouping for K and token-wise grouping for V.
Atom~\cite{zhao2024atom} and QServe~\cite{lin2025qserve} co-design 4-bit quantization formats and serving kernels, while GEAR~\cite{kang2024gear} adds low-rank and sparse error corrections.
\sys{} retains uniform affine quantization and changes how a Tensor Core kernel evaluates the quantized cache.
Nonlinear lookup formats and sparse outlier representations cannot be expressed as a single affine transform and require separate kernel support.

\subsection{Quantized attention kernels}

BitDecoding provides low-bit Tensor Core attention kernels across several GPU generations~\cite{du2026bitdecoding}.
On Hopper, it converts stored integer values and applies scales and offsets before the matrix products.
On Blackwell, it uses block-scaled FP4 instructions that restrict scales to the contraction axis, omit offsets, and quantize $Q$ and each $P$ tile to FP4.
TensorRT-LLM's XQA kernel folds a scalar INT8 or FP8 KV scale into the QK and PV accumulators~\cite{nvidia2026trtllm}.
QServe fuses low-bit KV dequantization into CUDA-core batched matrix-vector products~\cite{lin2025qserve}.
TurboAttention converts sub-8-bit values into INT8 operands for Tensor Core attention~\cite{kang2025turboattention}.
The SageAttention kernels quantize full-precision $Q$, $K$, and $V$ inputs for prefill and general attention workloads, including FP4 Tensor Core execution on Blackwell~\cite{zhang2025sageattention,zhang2025sageattention2,zhang2025sageattention3}.
SAW-INT4~\cite{jia2026saw} combines token-wise affine INT4 caches with a block-diagonal Hadamard rotation, while OSCAR~\cite{zhou2026oscar} combines a mixed INT2/BF16 cache with offline spectral rotations; both kernels reconstruct cached values before their matrix products.
\sys{} instead applies affine folding across independent token-wise or channel-wise K and V caches, folding group scales and non-zero offsets into the $Q$ or $P$ inputs or into group-local partial products while keeping $Q$ and $P$ in BF16 or FP8.
Our serving evaluation combines affine folding with OSCAR's cache policy and rotations (\Cref{sec:eval-serving-throughput}).

\subsection{Attention pipelines and kernel reparameterization}

FlashAttention-2 introduces work partitioning for tiled attention~\cite{dao2024flashattention2}, FlashAttention-3 overlaps producer and consumer warpgroups through warp specialization~\cite{shah2024flashattention3}, FlashInfer provides paged-cache attention kernels for serving engines~\cite{ye2025flashinfer}, and FlashAttention-4 studies how Blackwell's higher Tensor Core throughput shifts bottlenecks toward on-chip staging and non-matmul instructions~\cite{zadouri2026flashattention4}.
RepVGG~\cite{ding2021repvgg} folds training-time convolution branches into inference kernels, CODA~\cite{guo2026coda} rewrites Transformer operators into GEMM epilogues, and quantized GEMM kernels~\cite{zhao2024atom,lin2025qserve} rescale group partial sums or subtract zero-point activation sums in their epilogues.
\sys{} combines warp-specialized attention pipelining with scale-and-offset folding across both QK and PV under online softmax, where the eliminated per-element work grows with context length.

\subsection{Quantized inference architecture}

Oaken combines a KV quantization policy with custom hardware for online and offline compression~\cite{kim2025oaken}.
Weight-only kernels such as Marlin~\cite{frantar2025marlin,lin2025marlinw4a8}, FLUTE~\cite{guo2024flute}, and LUT-GEMM~\cite{park2024lutgemm} optimize single matrix products over quantized model weights.
\sys{} instead targets the QK and PV matrix products of decode attention, which are linked by online softmax, on unmodified Hopper and Blackwell Tensor Cores.

%% file: src/7-conclusion.tex
\section{Conclusion}
\label{sec:conclusion}

\sys{} accelerates decode attention over low-bit KV caches through affine folding, which moves each quantization group's shared scale to a matrix-product input or partial result and applies its offset as an output correction.
Its Hopper and Blackwell kernels construct Tensor Core KV operands from packed quantized values through bit manipulation, while affine folding preserves the attention output of the reconstructed cache in exact arithmetic.
On H100 and B200, \sys{} INT2 outperforms FlashInfer BF16 and TensorRT-LLM FP8 across all tested batch sizes and context lengths on the main benchmark grid.
Qwen3-8B serving experiments also show throughput gains of up to $170\%$ over baselines in settings with long contexts and high cache hit rates that align with popular agentic workloads.

%% file: src/8-appendix.tex
\appendix
\crefalias{section}{appendix}

\section{KV Packing and Dataflow}
\label{sec:operand-construction}

\Cref{fig:dataflow} illustrates how \sys{} packs INT2 K values and constructs Tensor Core operands on Hopper and Blackwell.
The same approach extends to INT2 V and INT4 K and V because cache packing arranges quantized values to match each Tensor Core operand layout, and the decode kernel constructs floating-point operands with shifts and masks.
Quantizing K or V along tokens or channels changes how scales and offsets are grouped and where affine folding applies them (\Cref{sec:design-fold}), whereas the integer packing below depends only on the stored bit width and matrix operand layout.

\subsection{Hopper: packing for register operands}

Hopper packs K and V according to the register fragments that each consumer thread supplies to \texttt{wgmma}.
For QK, the K operand has token-by-channel coordinates, as shown in \Cref{fig:dataflow-packing}.
For PV, the V operand has channel-by-token coordinates, so \sys{} packs V values according to the transposed V tile.
Both layouts let each consumer thread load contiguous packed words from shared memory and construct its register operands without gathering individual values.

Each 32-bit packed word places the two quantized values for a pair of BF16 operands in fields sixteen bits apart.
An INT2 word holds sixteen values, forming eight pairs; an INT4 word holds eight values, forming four pairs.
A shift aligns both fields with their BF16 mantissas, and a masked merge combines them with a fixed BF16 bit pattern.
The stored bit width sets the field spacing and the number of pairs per word.
K and V differ only in the logical coordinates of those pairs and share the same pairwise register construction.
Affine folding compensates for the resulting operand mapping through \Cref{eq:operand-transform}.

The Hopper dataflow in \Cref{fig:dataflow-threads} therefore also applies to INT2 V and INT4 K and V.
TMA copies the packed tiles from global to shared memory.
Consumer threads construct BF16 operands in registers, and all four warps of a consumer warpgroup collectively issue \texttt{wgmma}.
The QK product consumes K fragments, and the PV product consumes the transposed V fragments.

\subsection{Blackwell: packing for shared-memory operands}

The Blackwell kernel uses the same packed-word layout for K and V at each stored bit width.
Both tensors store channels of a token together, with the values interleaved within each word for direct operand construction.
The preparation warpgroup writes the constructed operands to the shared-memory layouts required by QK and PV.

\paragraph{INT2 K and V.}
Each 32-bit word holds sixteen integer values in eight four-bit groups.
Group $j$ places $q_j$ in its low two bits and $q_{j+8}$ in its high two bits, for $j=0,\ldots,7$.
For a packed word $w$, the kernel constructs two operand words:
\begin{align*}
  u_0 &= w \mathbin{\&} \texttt{0x33333333},\\
  u_1 &= (w \gg 2) \mathbin{\&} \texttt{0x33333333}.
\end{align*}
Each output holds eight E2M1 operands, with the unused bits of each four-bit operand set to zero.
The patterns for $q\in\{0,1,2,3\}$ represent $y\in\{0,0.5,1,1.5\}$, giving $y=q/2$.
Affine folding compensates with $s'=2s$ and $z'=z$.
The two masks and one shift construct sixteen operands without numeric conversion, accounting for the INT2 cost in \Cref{tab:operand-construction}.
This is the Blackwell construction shown for K in \Cref{fig:dataflow-packing}; V uses the same bit operations.

\paragraph{INT4 K and V.}
Each 32-bit word holds eight integer values in four bytes.
Byte $j$ places $q_j$ in its low four bits and $q_{j+4}$ in its high four bits, for $j=0,\ldots,3$.
The corresponding construction is
\begin{align*}
  u_0 &= w \mathbin{\&} \texttt{0x0F0F0F0F},\\
  u_1 &= (w \gg 4) \mathbin{\&} \texttt{0x0F0F0F0F}.
\end{align*}
Each output holds four E4M3 operands, with each integer in the low four bits of its byte.
For $q\in\{0,\ldots,15\}$, these patterns represent $y=q\cdot2^{-9}$.
Affine folding compensates with $s'=512s$ and $z'=z$.
The two masks and one shift construct eight operands, accounting for the INT4 cost in \Cref{tab:operand-construction}.
Marlin W4A8 uses the same INT4-to-E4M3 mapping for weights~\cite{lin2025marlinw4a8}; \sys{} applies it to unsigned KV values, whose construction needs no sign extraction.

Seven of the sixteen INT4 bit patterns represent subnormal E4M3 values, for which the PTX specification leaves matrix-operand behavior undefined~\cite{nvidia2026ptx}.
We test these subnormal values on B200 across both matrix-operand positions and both shared-memory layouts used by the kernel, and all matrix-instruction outputs match FP64 reference results.

For both stored bit widths, the Blackwell dataflow follows \Cref{fig:dataflow-threads}.
TMA copies packed K and V tiles to shared memory, the preparation warpgroup constructs the QK and PV operands in shared memory, and a separate issue thread launches \texttt{tcgen05}.
The operand format changes from E2M1 for INT2 to E4M3 for INT4, while the division of work across copying, operand preparation, and matrix-product issue remains the same.

\section{Quantization-Axis Combinations}
\label{sec:appendix-axes}

\input{figures/quant-axes}

The main evaluation uses token-wise K and token-wise V with group size $g=128$, which applies acc-fold to QK and operand-fold to PV (\Cref{sec:design-fold}).
To evaluate all four combinations of token-wise and channel-wise K and V, we benchmark \sys{} INT2 and INT4 on H100 and B200 across $B\in\{1,8\}$ and $KV\in\{32\mathrm{K},128\mathrm{K}\}$ at $H_Q=32$, $H_{KV}=8$, $D=128$, and $g=128$, following the timing methodology of \Cref{sec:eval} and verifying every output against an FP32 reference.

\Cref{fig:quant-axes} compares the decode speed of the three channel-wise K and V combinations against token-wise K and V ($1.00$) at the same stored bit width and shape.
On H100 (\Cref{fig:quant-axes-h100}), the channel-wise combinations run at $0.84$--$1.04\times$ the speed of token-wise K and V at INT2 and $0.91$--$1.07\times$ at INT4, corresponding to $2.08$--$4.00\times$ (INT2) and $1.75$--$3.35\times$ (INT4) speedups over FlashInfer BF16 across all four axis combinations.
On B200 (\Cref{fig:quant-axes-b200}), the single-axis channel-wise combinations (\emph{K-ch / V-tok} and \emph{K-tok / V-ch}) run at $0.77$--$0.87\times$ the speed of token-wise K and V at INT2 and $0.80$--$0.92\times$ at INT4, while channel-wise K and V (\emph{K-ch / V-ch}) runs at $0.57$--$0.73\times$ at INT2 and $0.67$--$0.79\times$ at INT4, corresponding to $1.50$--$2.74\times$ (INT2) and $1.50$--$2.36\times$ (INT4) speedups over FlashInfer BF16 across all four combinations.
Channel-wise K and channel-wise V read more scales and offsets and perform per-group work along the token dimension: operand-fold on channel-wise K scales $Q$ for each token group, while acc-fold on channel-wise V rescales group-local PV partial results across token groups.
The relative gap is larger on B200 than on H100 for two architectural reasons: first, the Blackwell kernel tiles the sequence at $256$ tokens rather than $64$, so $g=128$ splits each sequence tile into two sub-tile groups and runs both the $Q'$-plane builder and the per-group PV accumulator drain inside the correction warpgroup; second, Blackwell's \texttt{tcgen05} QK product consumes an FP8 E4M3 query operand rather than Hopper's BF16 operand, so each channel-wise K group must compute a per-row maximum and power-of-two scale to requantize $Q'$ to E4M3.
With KIVI-style channel-wise K and token-wise V~\cite{liu2024kivi,hooper2024kvquant}, \sys{} runs at $0.89$--$1.04\times$ of token-wise speed on H100 ($2.42$--$3.56\times$ over FlashInfer BF16) and $0.85$--$0.87\times$ on B200 ($2.11$--$2.38\times$ over FlashInfer BF16) at INT2, and $0.96$--$1.07\times$ on H100 ($2.03$--$3.31\times$ over FlashInfer BF16) and $0.87$--$0.92\times$ on B200 ($1.94$--$2.05\times$ over FlashInfer BF16) at INT4.

\section{Head-Ratio and Group-Size Trade-off}
\label{sec:appendix-head-scaling}

\input{figures/affine-op-count}
\input{figures/head-scaling}

When $h_q=H_Q/H_{KV}$ query heads share one KV head and each quantization group spans $g$ values, affine folding performs one scale multiplication and one offset multiply-add per group for each query head, or $2h_q/g$ scalar affine operations per stored value.
Reconstructing each cached value once with $sq+z$ and reusing it across the $h_q$ query heads costs one multiply-add per stored value.
\Cref{fig:affine-op-count} compares the scalar affine operations per value as $h_q$ and $g$ vary.
At $h_q=4$ and $g=128$, affine folding uses $\frac{2 \times 4}{128}$ operations per value, a $16\times$ reduction.
Break-even occurs at $g=2h_q$; because standard grouped-query attention models use group sizes $g \ge 32 > 2h_q$, affine folding requires up to $64\times$ fewer scalar affine operations than per-element reconstruction.

\Cref{fig:head-scaling} measures this trade-off empirically by varying $H_{KV}$ from 1 to 32 ($h_q$ from 32 down to 1) with $H_Q=32$, $D=128$, and $g=128$ at 128K context for batches 1 and 8.
At $H_{KV}=1$ ($h_q=32$), the smaller KV cache and higher $2h_q/g$ ratio limit the gain over FlashInfer BF16 ($0.91\times$ and $1.10\times$ on H100; $1.32\times$ and $0.77\times$ on B200).
As $H_{KV}$ grows, KV traffic increases while $h_q$ decreases, raising \sys{} INT2's speedup over FlashInfer BF16 to $3.89\times$ at batch 1 and $7.68\times$ at batch 8 on H100, and to $2.75\times$ and $3.12\times$ on B200 (\Cref{fig:head-scaling-b1,fig:head-scaling-b8}).

\clearpage

%% file: figures/quant-axes.tex
\begin{figure}[t]
  \centering
  \includegraphics[width=\columnwidth]{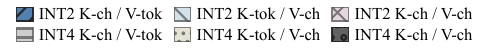}
  \par
  \subfloat[H100.\label{fig:quant-axes-h100}]{%
    \includegraphics[width=0.49\columnwidth]{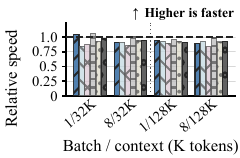}}
  \hfill
  \subfloat[B200.\label{fig:quant-axes-b200}]{%
    \includegraphics[width=0.49\columnwidth]{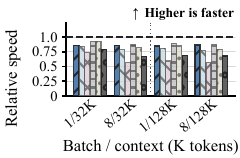}}
  \caption{Decode kernel speed across channel-wise K and V quantization combinations relative to token-wise K and V ($1.0$, dashed line) at the same stored bit width and shape ($H_Q=32$, $H_{KV}=8$, $D=128$, group size $g=128$).}
  \label{fig:quant-axes}
\end{figure}

%% file: figures/affine-op-count.tex
\begin{figure}[t]
  \centering
  \includegraphics[width=\columnwidth]{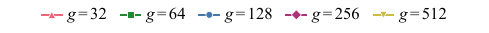}
  \par
  \subfloat[Token-wise quantization.\label{fig:affine-op-token}]{%
    \includegraphics[width=0.49\columnwidth]{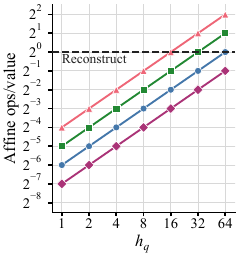}}
  \hfill
  \subfloat[Channel-wise quantization.\label{fig:affine-op-channel}]{%
    \includegraphics[width=0.49\columnwidth]{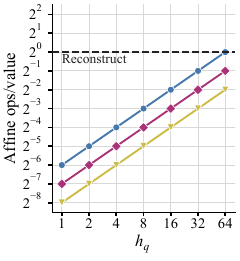}}
  \caption{Scalar scale and offset operations per stored K or V value for affine folding versus element-wise $sq+z$ reconstruction (dashed).}
  \label{fig:affine-op-count}
\end{figure}

%% file: figures/head-scaling.tex
\begin{figure}[t]
  \centering
  \captionsetup[subfloat]{farskip=2pt}
  \includegraphics[width=\columnwidth]{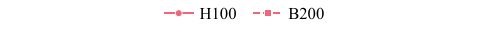}
  \par
  \subfloat[Batch 1.\label{fig:head-scaling-b1}]{%
    \includegraphics{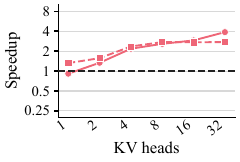}}
  \hfill
  \subfloat[Batch 8.\label{fig:head-scaling-b8}]{%
    \includegraphics{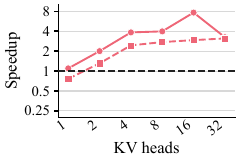}}
  \caption{\sys{} INT2 speedup over FlashInfer BF16 across $H_{KV}$ at 128K context ($H_Q=32$, $D=128$).}
  \label{fig:head-scaling}
\end{figure}

%% file: refs.bib
@inproceedings{banner2019posttraining,
  author       = {Ron Banner and
                  Yury Nahshan and
                  Daniel Soudry},
  editor       = {Hanna M. Wallach and
                  Hugo Larochelle and
                  Alina Beygelzimer and
                  Florence d'Alch{\'{e}}{-}Buc and
                  Emily B. Fox and
                  Roman Garnett},
  title        = {Post training 4-bit quantization of convolutional networks for rapid-deployment},
  booktitle    = {Advances in Neural Information Processing Systems 32: Annual Conference
                  on Neural Information Processing Systems 2019, NeurIPS 2019, December
                  8-14, 2019, Vancouver, BC, Canada},
  pages        = {7948--7956},
  year         = {2019},
  url          = {https://proceedings.neurips.cc/paper/2019/hash/c0a62e133894cdce435bcb4a5df1db2d-Abstract.html},
  bibsource    = {dblp computer science bibliography, https://dblp.org}
}

@inproceedings{yao2022zeroquant,
  author       = {Zhewei Yao and
                  Reza Yazdani Aminabadi and
                  Minjia Zhang and
                  Xiaoxia Wu and
                  Conglong Li and
                  Yuxiong He},
  editor       = {Sanmi Koyejo and
                  S. Mohamed and
                  A. Agarwal and
                  Danielle Belgrave and
                  K. Cho and
                  A. Oh},
  title        = {ZeroQuant: Efficient and Affordable Post-Training Quantization for
                  Large-Scale Transformers},
  booktitle    = {Advances in Neural Information Processing Systems 35: Annual Conference
                  on Neural Information Processing Systems 2022, NeurIPS 2022, New Orleans,
                  LA, USA, November 28 - December 9, 2022},
  year         = {2022},
  url          = {http://papers.nips.cc/paper\_files/paper/2022/hash/adf7fa39d65e2983d724ff7da57f00ac-Abstract-Conference.html},
  bibsource    = {dblp computer science bibliography, https://dblp.org}
}

@inproceedings{liu2024kivi,
  author       = {Zirui Liu and
                  Jiayi Yuan and
                  Hongye Jin and
                  Shaochen (Henry) Zhong and
                  Zhaozhuo Xu and
                  Vladimir Braverman and
                  Beidi Chen and
                  Xia Hu},
  editor       = {Ruslan Salakhutdinov and
                  Zico Kolter and
                  Katherine A. Heller and
                  Adrian Weller and
                  Nuria Oliver and
                  Jonathan Scarlett and
                  Felix Berkenkamp},
  title        = {{KIVI:} {A} Tuning-Free Asymmetric 2bit Quantization for {KV} Cache},
  booktitle    = {Forty-first International Conference on Machine Learning, {ICML} 2024,
                  Vienna, Austria, July 21-27, 2024},
  series       = {Proceedings of Machine Learning Research},
  volume       = {235},
  pages        = {32332--32344},
  publisher    = {{PMLR} / OpenReview.net},
  year         = {2024},
  url          = {https://proceedings.mlr.press/v235/liu24bz.html},
  bibsource    = {dblp computer science bibliography, https://dblp.org}
}

@inproceedings{hooper2024kvquant,
  author       = {Coleman Hooper and
                  Sehoon Kim and
                  Hiva Mohammadzadeh and
                  Michael W. Mahoney and
                  Yakun Sophia Shao and
                  Kurt Keutzer and
                  Amir Gholami},
  editor       = {Amir Globersons and
                  Lester Mackey and
                  Danielle Belgrave and
                  Angela Fan and
                  Ulrich Paquet and
                  Jakub M. Tomczak and
                  Cheng Zhang},
  title        = {KVQuant: Towards 10 Million Context Length {LLM} Inference with {KV}
                  Cache Quantization},
  booktitle    = {Advances in Neural Information Processing Systems 37: Annual Conference
                  on Neural Information Processing Systems 2024, NeurIPS 2024, Vancouver,
                  BC, Canada, December 10 - 15, 2024},
  year         = {2024},
  url          = {http://papers.nips.cc/paper\_files/paper/2024/hash/028fcbcf85435d39a40c4d61b42c99a4-Abstract-Conference.html},
  bibsource    = {dblp computer science bibliography, https://dblp.org}
}

@inproceedings{lin2025qserve,
  author       = {Yujun Lin and
                  Haotian Tang and
                  Shang Yang and
                  Zhekai Zhang and
                  Guangxuan Xiao and
                  Chuang Gan and
                  Song Han},
  editor       = {Matei Zaharia and
                  Gauri Joshi and
                  Yingyan (Celine) Lin},
  title        = {QServe: {W4A8KV4} Quantization and System Co-design for Efficient
                  {LLM} Serving},
  booktitle    = {Proceedings of the Eighth Conference on Machine Learning and Systems,
                  MLSys 2025, Santa Clara, CA, USA, May 12-15, 2025},
  publisher    = {OpenReview.net/mlsys.org},
  year         = {2025},
  url          = {https://openreview.net/forum?id=1FfmStySS1},
  bibsource    = {dblp computer science bibliography, https://dblp.org}
}

@inproceedings{zhao2024atom,
  author       = {Yilong Zhao and
                  Chien{-}Yu Lin and
                  Kan Zhu and
                  Zihao Ye and
                  Lequn Chen and
                  Size Zheng and
                  Luis Ceze and
                  Arvind Krishnamurthy and
                  Tianqi Chen and
                  Baris Kasikci},
  editor       = {Phillip B. Gibbons and
                  Gennady Pekhimenko and
                  Christopher De Sa},
  title        = {Atom: Low-Bit Quantization for Efficient and Accurate {LLM} Serving},
  booktitle    = {Proceedings of the Seventh Annual Conference on Machine Learning and
                  Systems, MLSys 2024, Santa Clara, CA, USA, May 13-16, 2024},
  publisher    = {mlsys.org},
  year         = {2024},
  url          = {https://proceedings.mlsys.org/paper\_files/paper/2024/hash/5edb57c05c81d04beb716ef1d542fe9e-Abstract-Conference.html},
  bibsource    = {dblp computer science bibliography, https://dblp.org}
}

@article{kang2024gear,
  author       = {Hao Kang and
                  Qingru Zhang and
                  Souvik Kundu and
                  Geonhwa Jeong and
                  Zaoxing Liu and
                  Tushar Krishna and
                  Tuo Zhao},
  title        = {{GEAR:} An Efficient {KV} Cache Compression Recipe for Near-Lossless
                  Generative Inference of {LLM}},
  journal      = {CoRR},
  volume       = {abs/2403.05527},
  year         = {2024},
  url          = {https://doi.org/10.48550/arXiv.2403.05527},
  doi          = {10.48550/ARXIV.2403.05527},
  eprinttype   = {arXiv},
  eprint       = {2403.05527},
  bibsource    = {dblp computer science bibliography, https://dblp.org}
}

@inproceedings{du2026bitdecoding,
  author       = {Dayou Du and
                  Shijie Cao and
                  Jianyi Cheng and
                  Luo Mai and
                  Ting Cao and
                  Mao Yang},
  title        = {BitDecoding: Unlocking Tensor Cores for Long-Context LLMs with Low-Bit
                  {KV} Cache},
  booktitle    = {{IEEE} International Symposium on High Performance Computer Architecture,
                  {HPCA} 2026, Sydney, Australia, January 31 - Feb. 4, 2026},
  pages        = {1--13},
  publisher    = {{IEEE}},
  year         = {2026},
  url          = {https://doi.org/10.1109/HPCA68181.2026.11408481},
  doi          = {10.1109/HPCA68181.2026.11408481},
  bibsource    = {dblp computer science bibliography, https://dblp.org}
}

@article{jia2026saw,
  author       = {Jinda Jia and
                  Jisen Li and
                  Zhongzhu Zhou and
                  Jung Hwan Heo and
                  Jue Wang and
                  Tri Dao and
                  Shuaiwen Leon Song and
                  Ben Athiwaratkun and
                  Chenfeng Xu and
                  Tianyi Zhang and
                  Xiaoxia Wu},
  title        = {{SAW-INT4:} System-Aware 4-Bit KV-Cache Quantization for Real-World
                  {LLM} Serving},
  journal      = {CoRR},
  volume       = {abs/2604.19157},
  year         = {2026},
  url          = {https://doi.org/10.48550/arXiv.2604.19157},
  doi          = {10.48550/ARXIV.2604.19157},
  eprinttype   = {arXiv},
  eprint       = {2604.19157},
  bibsource    = {dblp computer science bibliography, https://dblp.org}
}

@article{zhou2026oscar,
  author       = {Zhongzhu Zhou and
                  Donglin Zhuang and
                  Jisen Li and
                  Ziyan Chen and
                  Shuaiwen Leon Song and
                  Ben Athiwaratkun and
                  Xiaoxia Wu},
  title        = {{OSCAR:} Offline Spectral Covariance-Aware Rotation for 2-bit {KV}
                  Cache Quantization},
  journal      = {CoRR},
  volume       = {abs/2605.17757},
  year         = {2026},
  url          = {https://doi.org/10.48550/arXiv.2605.17757},
  doi          = {10.48550/ARXIV.2605.17757},
  eprinttype   = {arXiv},
  eprint       = {2605.17757},
  bibsource    = {dblp computer science bibliography, https://dblp.org}
}

@article{xia2025kitty,
  author       = {Haojun Xia and
                  Xiaoxia Wu and
                  Jisen Li and
                  Robert Wu and
                  Junxiong Wang and
                  Jue Wang and
                  Chenxi Li and
                  Aman Singhal and
                  Alay Dilipbhai Shah and
                  Alpay Ariyak and
                  Donglin Zhuang and
                  Zhongzhu Zhou and
                  Ben Athiwaratkun and
                  Zhen Zheng and
                  Shuaiwen Leon Song},
  title        = {Kitty: Accurate and Efficient 2-bit {KV} Cache Quantization with Dynamic
                  Channel-wise Precision Boost},
  journal      = {CoRR},
  volume       = {abs/2511.18643},
  year         = {2025},
  url          = {https://doi.org/10.48550/arXiv.2511.18643},
  doi          = {10.48550/ARXIV.2511.18643},
  eprinttype   = {arXiv},
  eprint       = {2511.18643},
  bibsource    = {dblp computer science bibliography, https://dblp.org}
}

@inproceedings{kang2025turboattention,
  author       = {Hao Kang and
                  Srikant Bharadwaj and
                  James Hensman and
                  Tushar Krishna and
                  Victor R{\"{u}}hle and
                  Saravan Rajmohan},
  editor       = {Matei Zaharia and
                  Gauri Joshi and
                  Yingyan (Celine) Lin},
  title        = {TurboAttention: Efficient attention approximation for high throughputs
                  llm},
  booktitle    = {Proceedings of the Eighth Conference on Machine Learning and Systems,
                  MLSys 2025, Santa Clara, CA, USA, May 12-15, 2025},
  publisher    = {OpenReview.net/mlsys.org},
  year         = {2025},
  url          = {https://openreview.net/forum?id=bPfpSO2zX4},
  bibsource    = {dblp computer science bibliography, https://dblp.org}
}

@inproceedings{zhang2025sageattention,
  author       = {Jintao Zhang and
                  Jia Wei and
                  Pengle Zhang and
                  Jun Zhu and
                  Jianfei Chen},
  title        = {SageAttention: Accurate 8-Bit Attention for Plug-and-play Inference
                  Acceleration},
  booktitle    = {The Thirteenth International Conference on Learning Representations,
                  {ICLR} 2025, Singapore, April 24-28, 2025},
  publisher    = {OpenReview.net},
  year         = {2025},
  url          = {https://openreview.net/forum?id=OL44KtasKc},
  bibsource    = {dblp computer science bibliography, https://dblp.org}
}

@inproceedings{zhang2025sageattention2,
  author       = {Jintao Zhang and
                  Haofeng Huang and
                  Pengle Zhang and
                  Jia Wei and
                  Jun Zhu and
                  Jianfei Chen},
  editor       = {Aarti Singh and
                  Maryam Fazel and
                  Daniel Hsu and
                  Simon Lacoste{-}Julien and
                  Felix Berkenkamp and
                  Tegan Maharaj and
                  Kiri Wagstaff and
                  Jerry Zhu},
  title        = {SageAttention2: Efficient Attention with Thorough Outlier Smoothing
                  and Per-thread {INT4} Quantization},
  booktitle    = {Forty-second International Conference on Machine Learning, {ICML}
                  2025, Vancouver, BC, Canada, July 13-19, 2025},
  series       = {Proceedings of Machine Learning Research},
  volume       = {267},
  publisher    = {{PMLR} / OpenReview.net},
  year         = {2025},
  url          = {https://proceedings.mlr.press/v267/zhang25ae.html},
  bibsource    = {dblp computer science bibliography, https://dblp.org}
}

@inproceedings{zhang2025sageattention3,
  author       = {Jintao Zhang and
                  Jia Wei and
                  Haoxu Wang and
                  Pengle Zhang and
                  Xiaoming Xu and
                  Haofeng Huang and
                  Kai Jiang and
                  Jianfei Chen and
                  Jun Zhu},
  editor       = {Danielle Belgrave and
                  Cheng Zhang and
                  Laura N. Montoya and
                  Hsuan{-}Tien Lin and
                  Razvan Pascanu and
                  Piotr Koniusz and
                  Marzyeh Ghassemi and
                  Nancy Chen and
                  Iv{\'{a}}n Vladimir Meza Ru{\'{\i}}z and
                  Arturo Loaiza{-}Bonilla},
  title        = {SageAttention3: Microscaling {FP4} Attention for Inference and An
                  Exploration of 8-Bit Training},
  booktitle    = {Advances in Neural Information Processing Systems 38: Annual Conference
                  on Neural Information Processing Systems 2025, NeurIPS 2025, San Diego,
                  CA, USA, December 2-7, 2025 / Mexico City, Mexico, November 30 - December
                  5, 2025},
  year         = {2025},
  url          = {http://papers.nips.cc/paper\_files/paper/2025/hash/4db397e0f760cc573c681e81a01a3dba-Abstract-Conference.html},
  bibsource    = {dblp computer science bibliography, https://dblp.org}
}

@inproceedings{frantar2025marlin,
  author       = {Elias Frantar and
                  Roberto L. Castro and
                  Jiale Chen and
                  Torsten Hoefler and
                  Dan Alistarh},
  title        = {{MARLIN:} Mixed-Precision Auto-Regressive Parallel Inference on Large
                  Language Models},
  booktitle    = {Proceedings of the 30th {ACM} {SIGPLAN} Annual Symposium on Principles
                  and Practice of Parallel Programming, PPoPP 2025, Las Vegas, NV, USA,
                  March 1-5, 2025},
  pages        = {239--251},
  publisher    = {{ACM}},
  year         = {2025},
  url          = {https://doi.org/10.1145/3710848.3710871},
  doi          = {10.1145/3710848.3710871},
  bibsource    = {dblp computer science bibliography, https://dblp.org}
}

@inproceedings{guo2024flute,
  author       = {Han Guo and
                  William Brandon and
                  Radostin Cholakov and
                  Jonathan Ragan{-}Kelley and
                  Eric P. Xing and
                  Yoon Kim},
  editor       = {Yaser Al{-}Onaizan and
                  Mohit Bansal and
                  Yun{-}Nung Chen},
  title        = {Fast Matrix Multiplications for Lookup Table-Quantized LLMs},
  booktitle    = {Findings of the Association for Computational Linguistics: {EMNLP}
                  2024, Miami, Florida, USA, November 12-16, 2024},
  series       = {Findings of {ACL}},
  volume       = {{EMNLP} 2024},
  pages        = {12419--12433},
  publisher    = {Association for Computational Linguistics},
  year         = {2024},
  url          = {https://doi.org/10.18653/v1/2024.findings-emnlp.724},
  doi          = {10.18653/V1/2024.FINDINGS-EMNLP.724},
  bibsource    = {dblp computer science bibliography, https://dblp.org}
}

@inproceedings{park2024lutgemm,
  author       = {Gunho Park and
                  Baeseong Park and
                  Minsub Kim and
                  Sungjae Lee and
                  Jeonghoon Kim and
                  Beomseok Kwon and
                  Se Jung Kwon and
                  Byeongwook Kim and
                  Youngjoo Lee and
                  Dongsoo Lee},
  title        = {{LUT-GEMM:} Quantized Matrix Multiplication based on LUTs for Efficient
                  Inference in Large-Scale Generative Language Models},
  booktitle    = {The Twelfth International Conference on Learning Representations,
                  {ICLR} 2024, Vienna, Austria, May 7-11, 2024},
  publisher    = {OpenReview.net},
  year         = {2024},
  url          = {https://openreview.net/forum?id=gLARhFLE0F},
  bibsource    = {dblp computer science bibliography, https://dblp.org}
}

@article{guo2026coda,
  author       = {Han Guo and
                  Jack Zhang and
                  Arjun Menon and
                  Driss Guessous and
                  Vijay Thakkar and
                  Yoon Kim and
                  Tri Dao},
  title        = {{CODA:} Rewriting Transformer Blocks as GEMM-Epilogue Programs},
  journal      = {CoRR},
  volume       = {abs/2605.19269},
  year         = {2026},
  url          = {https://doi.org/10.48550/arXiv.2605.19269},
  doi          = {10.48550/ARXIV.2605.19269},
  eprinttype   = {arXiv},
  eprint       = {2605.19269},
  bibsource    = {dblp computer science bibliography, https://dblp.org}
}

@inproceedings{ding2021repvgg,
  author       = {Xiaohan Ding and
                  Xiangyu Zhang and
                  Ningning Ma and
                  Jungong Han and
                  Guiguang Ding and
                  Jian Sun},
  title        = {RepVGG: Making VGG-Style ConvNets Great Again},
  booktitle    = {{IEEE} Conference on Computer Vision and Pattern Recognition, {CVPR}
                  2021, virtual, June 19-25, 2021},
  pages        = {13733--13742},
  publisher    = {Computer Vision Foundation / {IEEE}},
  year         = {2021},
  url          = {https://openaccess.thecvf.com/content/CVPR2021/html/Ding\_RepVGG\_Making\_VGG-Style\_ConvNets\_Great\_Again\_CVPR\_2021\_paper.html},
  doi          = {10.1109/CVPR46437.2021.01352},
  bibsource    = {dblp computer science bibliography, https://dblp.org}
}

@inproceedings{dao2022flashattention,
  author       = {Tri Dao and
                  Daniel Y. Fu and
                  Stefano Ermon and
                  Atri Rudra and
                  Christopher R{\'{e}}},
  editor       = {Sanmi Koyejo and
                  S. Mohamed and
                  A. Agarwal and
                  Danielle Belgrave and
                  K. Cho and
                  A. Oh},
  title        = {FlashAttention: Fast and Memory-Efficient Exact Attention with IO-Awareness},
  booktitle    = {Advances in Neural Information Processing Systems 35: Annual Conference
                  on Neural Information Processing Systems 2022, NeurIPS 2022, New Orleans,
                  LA, USA, November 28 - December 9, 2022},
  year         = {2022},
  url          = {http://papers.nips.cc/paper\_files/paper/2022/hash/67d57c32e20fd0a7a302cb81d36e40d5-Abstract-Conference.html},
  bibsource    = {dblp computer science bibliography, https://dblp.org}
}

@inproceedings{dao2024flashattention2,
  author       = {Tri Dao},
  title        = {FlashAttention-2: Faster Attention with Better Parallelism and Work
                  Partitioning},
  booktitle    = {The Twelfth International Conference on Learning Representations,
                  {ICLR} 2024, Vienna, Austria, May 7-11, 2024},
  publisher    = {OpenReview.net},
  year         = {2024},
  url          = {https://openreview.net/forum?id=mZn2Xyh9Ec},
  bibsource    = {dblp computer science bibliography, https://dblp.org}
}

@inproceedings{shah2024flashattention3,
  author       = {Jay Shah and
                  Ganesh Bikshandi and
                  Ying Zhang and
                  Vijay Thakkar and
                  Pradeep Ramani and
                  Tri Dao},
  editor       = {Amir Globersons and
                  Lester Mackey and
                  Danielle Belgrave and
                  Angela Fan and
                  Ulrich Paquet and
                  Jakub M. Tomczak and
                  Cheng Zhang},
  title        = {FlashAttention-3: Fast and Accurate Attention with Asynchrony and
                  Low-precision},
  booktitle    = {Advances in Neural Information Processing Systems 37: Annual Conference
                  on Neural Information Processing Systems 2024, NeurIPS 2024, Vancouver,
                  BC, Canada, December 10 - 15, 2024},
  year         = {2024},
  url          = {http://papers.nips.cc/paper\_files/paper/2024/hash/7ede97c3e082c6df10a8d6103a2eebd2-Abstract-Conference.html},
  bibsource    = {dblp computer science bibliography, https://dblp.org}
}

@article{zadouri2026flashattention4,
  author       = {Ted Zadouri and
                  Markus Hoehnerbach and
                  Jay Shah and
                  Timmy Liu and
                  Vijay Thakkar and
                  Tri Dao},
  title        = {FlashAttention-4: Algorithm and Kernel Pipelining Co-Design for Asymmetric
                  Hardware Scaling},
  journal      = {CoRR},
  volume       = {abs/2603.05451},
  year         = {2026},
  url          = {https://doi.org/10.48550/arXiv.2603.05451},
  doi          = {10.48550/ARXIV.2603.05451},
  eprinttype   = {arXiv},
  eprint       = {2603.05451},
  bibsource    = {dblp computer science bibliography, https://dblp.org}
}

@misc{dao2023flashdecoding,
  author       = {Tri Dao and Daniel Haziza and Francisco Massa and Grigory Sizov},
  title        = {{Flash-Decoding} for Long-Context Inference},
  howpublished = {{PyTorch} Blog},
  year         = {2023},
  url          = {https://pytorch.org/blog/flash-decoding/},
  note         = {Accessed 2026-08-14},
}

@inproceedings{ye2025flashinfer,
  author       = {Zihao Ye and
                  Lequn Chen and
                  Ruihang Lai and
                  Wuwei Lin and
                  Yineng Zhang and
                  Stephanie Wang and
                  Tianqi Chen and
                  Baris Kasikci and
                  Vinod Grover and
                  Arvind Krishnamurthy and
                  Luis Ceze},
  editor       = {Matei Zaharia and
                  Gauri Joshi and
                  Yingyan (Celine) Lin},
  title        = {FlashInfer: Efficient and Customizable Attention Engine for {LLM}
                  Inference Serving},
  booktitle    = {Proceedings of the Eighth Conference on Machine Learning and Systems,
                  MLSys 2025, Santa Clara, CA, USA, May 12-15, 2025},
  publisher    = {OpenReview.net/mlsys.org},
  year         = {2025},
  url          = {https://openreview.net/forum?id=RXPofAsL8F},
  bibsource    = {dblp computer science bibliography, https://dblp.org}
}

@inproceedings{zheng2024sglang,
  author       = {Lianmin Zheng and
                  Liangsheng Yin and
                  Zhiqiang Xie and
                  Chuyue Sun and
                  Jeff Huang and
                  Cody Hao Yu and
                  Shiyi Cao and
                  Christos Kozyrakis and
                  Ion Stoica and
                  Joseph E. Gonzalez and
                  Clark W. Barrett and
                  Ying Sheng},
  editor       = {Amir Globersons and
                  Lester Mackey and
                  Danielle Belgrave and
                  Angela Fan and
                  Ulrich Paquet and
                  Jakub M. Tomczak and
                  Cheng Zhang},
  title        = {SGLang: Efficient Execution of Structured Language Model Programs},
  booktitle    = {Advances in Neural Information Processing Systems 37: Annual Conference
                  on Neural Information Processing Systems 2024, NeurIPS 2024, Vancouver,
                  BC, Canada, December 10 - 15, 2024},
  year         = {2024},
  url          = {http://papers.nips.cc/paper\_files/paper/2024/hash/724be4472168f31ba1c9ac630f15dec8-Abstract-Conference.html},
  bibsource    = {dblp computer science bibliography, https://dblp.org}
}

@inproceedings{kim2025oaken,
  author       = {Minsu Kim and
                  Seongmin Hong and
                  Ryeowook Ko and
                  Soongyu Choi and
                  Hunjong Lee and
                  Junsoo Kim and
                  Joo{-}Young Kim and
                  Jongse Park},
  title        = {Oaken: Fast and Efficient {LLM} Serving with Online-Offline Hybrid
                  {KV} Cache Quantization},
  booktitle    = {Proceedings of the 52nd Annual International Symposium on Computer
                  Architecture, {ISCA} 2025, Tokyo, Japan, June 21-25, 2025},
  pages        = {482--497},
  publisher    = {{ACM}},
  year         = {2025},
  url          = {https://doi.org/10.1145/3695053.3731019},
  doi          = {10.1145/3695053.3731019},
  bibsource    = {dblp computer science bibliography, https://dblp.org}
}

@manual{nvidia2026ptx,
  author       = {{NVIDIA Corporation}},
  title        = {Parallel Thread Execution {ISA}, Version 9.4},
  organization = {NVIDIA Corporation},
  year         = {2026},
  url          = {https://docs.nvidia.com/cuda/parallel-thread-execution/index.html},
  note         = {Accessed 2026-10-29},
}

@manual{nvidia2026cudaguide,
  author       = {{NVIDIA Corporation}},
  title        = {{CUDA C++} Programming Guide},
  organization = {NVIDIA Corporation},
  year         = {2026},
  url          = {https://docs.nvidia.com/cuda/cuda-c-programming-guide/},
  note         = {Accessed 2026-08-20},
}

@manual{nvidia2026cudabpg,
  author       = {{NVIDIA Corporation}},
  title        = {{CUDA C++} Best Practices Guide},
  organization = {NVIDIA Corporation},
  year         = {2026},
  url          = {https://docs.nvidia.com/cuda/cuda-c-best-practices-guide/index.html\#arithmetic-instructions-throughput-native-arithmetic-instructions},
  note         = {Accessed 2026-07-30},
}

@manual{nvidia2023gpuperformance,
  author       = {{NVIDIA Corporation}},
  title        = {{GPU} Performance Background User's Guide},
  organization = {NVIDIA Corporation},
  year         = {2023},
  url          = {https://docs.nvidia.com/deeplearning/performance/dl-performance-gpu-background/index.html},
  note         = {Accessed 2026-08-12},
}

@manual{nvidia2025h100spec,
  author       = {{NVIDIA Corporation}},
  title        = {{NVIDIA H100} Tensor Core {GPU}},
  organization = {NVIDIA Corporation},
  year         = {2025},
  url          = {https://www.nvidia.com/en-us/data-center/h100/},
  note         = {Accessed 2026-08-05},
}

@manual{nvidia2025hgxb200,
  author       = {{NVIDIA Corporation}},
  title        = {{NVIDIA HGX} Platform},
  organization = {NVIDIA Corporation},
  year         = {2025},
  url          = {https://www.nvidia.com/en-us/data-center/hgx/},
  note         = {Accessed 2026-08-05},
}

@manual{nvidia2025dgxb200,
  author       = {{NVIDIA Corporation}},
  title        = {{NVIDIA DGX B200} Datasheet},
  organization = {NVIDIA Corporation},
  year         = {2025},
  url          = {https://resources.nvidia.com/en-us-dgx-systems/dgx-b200-datasheet},
  note         = {Accessed 2026-08-05},
}

@manual{nvidia2026rubin,
  author       = {{NVIDIA Corporation}},
  title        = {{NVIDIA Vera Rubin NVL72}},
  organization = {NVIDIA Corporation},
  year         = {2026},
  url          = {https://www.nvidia.com/en-us/data-center/vera-rubin-nvl72/},
  note         = {Accessed 2026-08-05},
}

@article{kim2022elephants,
  author       = {Young Jin Kim and
                  Rawn Henry and
                  Raffy Fahim and
                  Hany Hassan Awadalla},
  title        = {Who Says Elephants Can't Run: Bringing Large Scale MoE Models into
                  Cloud Scale Production},
  journal      = {CoRR},
  volume       = {abs/2211.10017},
  year         = {2022},
  url          = {https://doi.org/10.48550/arXiv.2211.10017},
  doi          = {10.48550/ARXIV.2211.10017},
  eprinttype   = {arXiv},
  eprint       = {2211.10017},
  bibsource    = {dblp computer science bibliography, https://dblp.org}
}

@misc{lin2025marlinw4a8,
  author       = {Jinzhen Lin},
  title        = {Add {W4A8} support for the {Marlin} kernel},
  howpublished = {{vLLM} pull request \#24722},
  year         = {2025},
  url          = {https://github.com/vllm-project/vllm/pull/24722},
  note         = {Accessed 2026-08-19},
}

@manual{nvidia2026trtllm,
  author       = {{NVIDIA Corporation}},
  title        = {{TensorRT-LLM}},
  organization = {NVIDIA Corporation},
  year         = {2026},
  url          = {https://github.com/NVIDIA/TensorRT-LLM},
  note         = {Accessed 2026-08-19},
}

@article{rein2023gpqa,
  author       = {David Rein and
                  Betty Li Hou and
                  Asa Cooper Stickland and
                  Jackson Petty and
                  Richard Yuanzhe Pang and
                  Julien Dirani and
                  Julian Michael and
                  Samuel R. Bowman},
  title        = {{GPQA:} {A} Graduate-Level Google-Proof Q{\&}A Benchmark},
  journal      = {CoRR},
  volume       = {abs/2311.12022},
  year         = {2023},
  url          = {https://doi.org/10.48550/arXiv.2311.12022},
  doi          = {10.48550/ARXIV.2311.12022},
  eprinttype   = {arXiv},
  eprint       = {2311.12022},
  bibsource    = {dblp computer science bibliography, https://dblp.org}
}
